 \documentclass[pmlr,twocolumn,10pt]{jmlr} 

\usepackage{siunitx}

\usepackage[switch]{lineno}
\usepackage{graphicx}%
\usepackage{multirow}%
\usepackage{amsmath,amssymb,amsfonts}%
\usepackage{mathrsfs}%
\usepackage[title]{appendix}%
\usepackage{xcolor}%
\usepackage{textcomp}%
\usepackage{manyfoot}%
\usepackage{booktabs}%
\usepackage{algorithm}%
\usepackage{algorithmicx}%
\usepackage{algpseudocode}%
\usepackage{listings}%
\usepackage{makecell}

\usepackage{xurl}
\usepackage{hyperref}

\usepackage{tikz}
\usepackage{subcaption}
\usepackage{caption}
\usetikzlibrary{arrows.meta,positioning,shapes.geometric,shapes.symbols,calc,fit}
\usepackage{multirow}
\usepackage{contour}

\usepackage{bm}

\newcommand{\equal}[1]{{\hypersetup{linkcolor=black}\thanks{#1}}}

\theorembodyfont{\upshape}
\theoremheaderfont{\scshape}
\theorempostheader{:}
\theoremsep{\newline}

\definecolor{myorange}{HTML}{E66101}

\jmlrvolume{297}
\jmlryear{2025}
\jmlrworkshop{Machine Learning for Health (ML4H) 2025} 

 \title[Valid Generative Survival Trial Data]{Toward Valid Generative Clinical Trial Data with Survival Endpoints}

\author{%
\Name{Perrine Chassat}\equal{These authors contributed equally} \Email{perrine.chassat@inria.fr}\\
\addr Inria, Université Paris Cité, Inserm, HeKA, F-75015 Paris, France
\AND
\Name{{Van Tuan} Nguyen}\footnotemark[1] \Email{van-tuan.nguyen@inria.fr}\\
\addr Inria, Université Paris Cité, Inserm, HeKA, F-75015 Paris, France
\AND
\Name{Lucas Ducrot}
\Email{lucas.ducrot@univ-rouen.fr}\\
\addr Inria, Université Paris Cité, Inserm, HeKA, F-75015 Paris, France
\AND
\Name{Emilie Lanoy}
\Email{emilie.lanoy@aphp.fr}\\
\addr AP-HP, Hôpital Européen Georges Pompidou, Unité de Recherche Clinique, APHP Centre, Paris, France. \\
\addr Inserm, Centre d'Investigation Clinique 1418 (CIC1418) Epidémiologie Clinique, Paris, France.
\AND
\Name{Agathe Guilloux} \Email{agathe.guilloux@inria.fr}\\
\addr Inria, Université Paris Cité, Inserm, HeKA, F-75015 Paris, France
}

\begin{document}

\maketitle

\begin{abstract}
Clinical trials face mounting challenges: fragmented patient populations, slow enrollment, and unsustainable costs, particularly for late phase trials in oncology and rare diseases. While external control arms built from real-world data have been explored, a promising alternative is the generation of synthetic control arms using generative AI. A central challenge is the generation of time-to-event outcomes, which constitute primary endpoints in oncology and rare disease trials, but are difficult to model under censoring and small sample sizes. Existing generative approaches, largely GAN-based, are data-hungry, unstable, and rely on strong assumptions such as independent censoring.
We introduce a variational autoencoder (VAE) that jointly generates mixed-type covariates and survival outcomes within a unified latent variable framework, without assuming independent censoring. Across synthetic and real trial datasets, we evaluate our model in two realistic scenarios: (i) data sharing under privacy constraints, where synthetic controls substitute for original data, and (ii) control-arm augmentation, where synthetic patients mitigate imbalances between treated and control groups. Our method outperforms GAN baselines on fidelity, utility, and privacy metrics, while revealing systematic miscalibration of type I error and power. We propose a post-generation selection procedure that improves calibration, highlighting both progress and open challenges for generative survival modeling.
\end{abstract}


\paragraph*{Data and Code Availability} We used phase III clinical trial datasets in our experiments. We accessed the publicly available AIDS Clinical Trials Group (ACTG) 320 study via \href{https://hivdb.stanford.edu/pages/clinicalStudyData/ACTG320.html}{Stanford University HIV drug resistance database}. The other 3 datasets - trials NCT00119613, NCT00113763, NCT00339183 - are accessible after registration at the \href{https://data.projectdatasphere.org}{Project Data Sphere}.
The code is publicly available at \href{https://github.com/aguilloux/survgen-clinical-trials}{https://github.com/aguilloux/survgen-clinical-trials}.

\paragraph*{Institutional Review Board (IRB)} This research does not require IRB approval. 


\section{Introduction}
\label{sec:intro}

Synthetic clinical data generation is an emerging frontier in machine learning for health, with applications ranging from privacy-preserving data sharing to benchmarking and clinical trial simulation. The clinical trial context is particularly compelling: as precision medicine fragments patient populations into increasingly fine strata (e.g., molecular subgroups), late Phase II-III randomized trials face slow enrollment, early discontinuation before reaching planned sample size, and in consequence escalating costs. These issues are most acute in oncology and rare diseases, where limited populations make traditional randomized designs difficult to execute. Synthetic data has therefore been proposed as a way to alleviate these barriers—by enabling external validation, supporting meta-analyses, or augmenting underpowered control arms.

Several approaches exist for generating patient-level data. Those based on mechanistic models provide interpretable scenarios grounded in biological knowledge \citep{carlier2018silico,delobel2023overcoming,wang2024virtual}. More recently, data-driven generative models have been applied to tabular clinical datasets, with methods such as TGAN, TVAE \citep{Xu2019}, and CTAB-GAN \citep{zhao2021ctab} producing synthetic samples that preserve statistical properties while protecting privacy \citep{d2023synthetic,petrakos2025framework}.

A central open challenge is the generation of time-to-event outcomes, frequently defined as the primary endpoints in most late-phase trials. Existing extensions of GANs to this setting, including SurvivalGAN \citep{norcliffe2023survivalgan}, have shown initial promise \citep{d2023synthetic,eckardt2024mimicking,akiya2024comparison,el2025augmenting,elvatun2025synthetic}, but face important limitations: they require large training datasets \citep{wang2023enhancing}, suffer from instability \citep{thanh2020catastrophic}, and assume independent random censoring, which is rarely realistic.

\paragraph{Contributions and Paper Outline}  
Our contributions are threefold:  
(i) We introduce a novel variational autoencoder (VAE) that jointly generates mixed-type covariates and survival outcomes within a unified latent variable framework, without assuming independent censoring.  
(ii) We propose a calibration-oriented evaluation framework, going beyond fidelity, utility, and privacy to assess type I error and power in downstream survival analyses.  
(iii) We investigate two realistic trial scenarios—data sharing under privacy constraints and control-arm augmentation—and show that while our HI-VAE variants outperform GAN and VAE baselines on classical metrics, all models fail to reproduce nominal type I error and power without additional safeguards. We propose a post-generation selection procedure that improves calibration, partially restoring statistical validity. Beyond its methodological contribution, this work also aims to inform current regulatory initiatives at the European Medicines Agency (EMA\footnote{https://www.ema.europa.eu/en/human-regulatory-overview/research-development/innovation-task-force-briefing-meetings}) and Food and Drug Administration (FDA\footnote{https://www.fda.gov/about-fda/cder-center-clinical-trial-innovation-c3ti/c3ti-areas-interest}) on the evaluation and potential integration of AI-based and model-based evidence in clinical development.

The remainder of the paper is organized as follows. Section~\ref{sec:Method} details our model. Section~\ref{sec:experiments} presents the experimental setup, including datasets, baselines, and evaluation metrics. Section~\ref{sec:results} reports our findings, and discussion. 

\section{Method}
\label{sec:Method}
Formally, we consider that the observed tabular data from a clinical trial can be represented as
\[\mathcal{D} = \{(x_i, t_i, \delta_i)\}_{i=1}^N,
\]
where $x_i \in \mathbb{R}^d$ denotes a vector of covariates for subject $i$; unless otherwise specified, this vector includes the treatment assignment variable $e_i \in \{0,1\}$, which indicates whether the individual received the treatment (1) or was assigned to the control group (0). The variable $t_i > 0$ represents the observed time, corresponding either to an event time (primary endpoint) or a censoring time, and $\delta_i \in \{0,1\}$ is the event indicator (1 if the event occurred, 0 if censored). Finally, $N = N^T + N^C$ denotes the total number of trial participants, with $N^T$ (respectively $N^C$) individuals assigned to the treatment (respectively control) arm.

To generate synthetic clinical trial data, we extend the HI-VAE (Heterogeneous and Incomplete Variational Autoencoder) of~\cite{Nazabal2020} to explicitly model censored survival times. The HI-VAE is a VAE-based model designed to handle missing data, in which the Gaussian prior is replaced by a mixture of Gaussians, allowing it to capture greater heterogeneity in complex clinical data. This choice is further motivated by the advantages of VAE-based formulations for tabular data generation over GAN-based models in terms of training stability, data efficiency, and interpretable latent representations \citep{Xu2019}.

More specifically, we consider a hierarchical model with three latent variables for each subject $i$.
The first two variables jointly define a Gaussian mixture in the latent space: a one-hot vector $\mathbf{s}_i \in \{0,1\}^L$ indicates the mixture component and is drawn from a categorical prior with uniform mixing probabilities, and a continuous latent variable $\mathbf{z}_i \in \mathbb{R}^K$ is drawn conditionally on $\mathbf{s}_i$. Their priors are given by
\begin{align*}
&p(\mathbf{s}_i) = \mathrm{Cat}(\mathbf{s}_i \mid \boldsymbol{\pi}), 
\quad \boldsymbol{\pi}_\ell = 1/L, \\
&p(\mathbf{z}_i \mid \mathbf{s}_i) = \mathcal{N}\bigl(\mathbf{z}_i \mid \boldsymbol{\mu}_p(\mathbf{s}_i), \mathrm{Id}\bigr),
\end{align*}
where $\{\boldsymbol{\mu}_\ell\}_{\ell=1}^L$ denote the component means. 
Finally, an intermediate representation $\mathbf{y}_i = g(\mathbf{z}_i) \in \mathbb{R}^H$ is obtained through a neural network $g$, which maps $\mathbf{z}_i$ to a homogeneous latent space that captures dependencies among heterogeneous attributes.
The generative process is parameterized by the joint likelihood of the observed data $(x_i, t_i, \delta_i)$, factorized as:
\[
p(x_i, t_i, \delta_i \mid \mathbf{z}_i,\mathbf{s}_i) = \prod_{j=1}^d p\bigl(x_i^j \mid h_j^i\bigr)  p(t_i, \delta_i \mid \mathbf{z}_i,\mathbf{s}_i) 
\]
where $h_j^i = h_j(\mathbf{y}_i, \mathbf{s}_i)$, $h_j$ is neural network, which outputs the parameters of the feature-specific conditional density  $p(x_i^j \mid h_j(\mathbf{y}_i, \mathbf{s}_i))$ determined by the type of the $j$-th variable. 
Common choices include, but are not limited to, the Gaussian distribution for continuous variables, with
\[ h_j(\mathbf{y}_i, \mathbf{s}_i) = \big(\mu_j(\mathbf{y}_i, \mathbf{s}_i), \sigma_j^2(\mathbf{y}_i, \mathbf{s}_i)\big), \] 
the log-normal or Poisson distributions for strictly positive or count data, and the multinomial-logit distribution for categorical variables (for more details, we refer the reader to Appendix~\ref{apd:model_detail}). 

For the part of the distribution associated with the time-to-event outcomes $p(t_i, \delta_i \mid \mathbf{z}_i,\mathbf{s}_i)$, we chose to parameterize the complete density (rather than only the distribution of the event times) as
\begin{align*}       &\Bigl[p\bigl(t_i \mid \eta_T(\mathbf{y}_i, \mathbf{s}_i)\bigr) \, \bar{P}\bigl(t_i \mid \eta_C(\mathbf{y}_i, \mathbf{s}_i)\bigr)\Bigr]^{\delta_i} \\ &\Bigl[\bar{P}\bigl(t_i \mid \eta_T(\mathbf{y}_i, \mathbf{s}_i)\bigr) \, p\bigl(t_i \mid \eta_C(\mathbf{y}_i, \mathbf{s}_i)\bigr)\Bigr]^{1 - \delta_i}, 
\end{align*}
where $p$ denotes a density and $\bar{P}$ the associated survival function. These are parameterized by neural networks: $\eta_T(\mathbf{y}_i, \mathbf{s}_i)$ for the event time and $\eta_C(\mathbf{y}_i, \mathbf{s}_i)$ for the censoring time. 
In practice, several choices are possible for $p$. In our implementation, it is modeled using either a Weibull distribution or a piecewise-constant density model.  We refer to these two variants as \texttt{HI-VAE\_Weibull} and \texttt{HI-VAE\_piecewise}, respectively. The detailed formulations of these two choices are provided in Appendix~\ref{apd:model_detail}.


For recognition models, we define the variational approximate posteriors as
$q(\mathbf{s}_i \mid x_i, t_i, \delta_i) = \mathrm{Cat}(\mathbf{s}_i \mid \boldsymbol{\pi}(x_i, t_i, \delta_i)), \quad
q(\mathbf{z}_i \mid x_i, t_i, \delta_i, \mathbf{s}_i) = \mathcal{N}(\mathbf{z}_i \mid \boldsymbol{\mu}_q(x_i, t_i, \delta_i, \mathbf{s}_i), \boldsymbol{\Sigma}_q(x_i, t_i, \delta_i, \mathbf{s}_i))$
where $\boldsymbol{\pi}$, $\boldsymbol{\mu}_q$ and $\boldsymbol{\Sigma}_q$ are parametrized by independent neural networks.

The Evidence Lower Bound (ELBO) to be optimized from the data writes 
\begin{align}
\label{eq:ELBO}
 &\log p(\mathcal{D}) = \log p \bigl( x_i, t_i, \delta_i, i=1, \ldots,N\bigr) \notag \\
 & \geq \mathcal{L}_{\mathrm{HI\text{-}VAE}} = 
  \sum_{i=1}^N \mathcal{L}_{\mathrm{HI\text{-}VAE}}^i
\end{align}
and \begin{align*}
&\mathcal{L}_{\mathrm{HI\text{-}VAE}}^i=\mathbb{E}_{q(\mathbf{z}_i,\mathbf{s}_i|x_i, t_i, \delta_i)}\bigl[p(x_i, t_i, \delta_i \mid \mathbf{z}_i,\mathbf{s}_i)\bigr]\notag 
 \\& - \mathbb{E}_{\{q(\mathbf{s}_i|x_i, t_i, \delta_i)\}}\bigl[\mathrm{KL}(q(\mathbf{z}_i|x_i, t_i, \delta_i,\mathbf{s}_i)|p(\mathbf{z}_i|\mathbf{s}_i))\bigr]\notag \\
 & - \mathrm{KL}(q(\mathbf{s}_i|x_i, t_i, \delta_i)|p(\mathbf{s}_i)),
\end{align*}
here $\mathrm{KL}$ stands for the Kullback–Leibler divergence. 

Synthetic samples are then generated through posterior sampling. Specifically, each original data point $(x_i, t_i, \delta_i)$ is first encoded into latent representations using the variational encoders $q(\mathbf{s}_i \mid x_i, t_i, \delta_i)$ and $q(\mathbf{z}_i \mid x_i, t_i, \delta_i, \mathbf{s}_i)$. Differentiable samples are then drawn from these learned posteriors using the Gumbel–Softmax trick \citep{Li2019LearningLS} for $\mathbf{s}_i$, and the Gaussian reparameterization trick \citep{Kingma2013AutoEncodingVB} for $\mathbf{z}_i \mid \mathbf{s}_i$.  Finally, the sampled latent variables are passed through the decoder $p(x_i, t_i, \delta_i \mid \mathbf{z}_i, \mathbf{s}_i)$ to generate synthetic data. This sampling procedure ensures that the generated samples reflect the statistical structure learned from the original dataset \citep{Nazabal2020}.

\section{Experiments}\label{sec:experiments}

 We frame our experiments in the standard late phase randomized clinical trial setting. A typical analysis proceeds in two steps. First, the performance of randomization in balancing patients characteristics across experimental and control treatment arms is assessed as recommended in established reporting guidelines~\citep{moher2010consort}. Second, the experimental treatment effect on survival outcomes is evaluated, most commonly using a log-rank test to compare survival distributions and/or by estimating the hazard ratio with a Cox proportional hazards model \citep{fleming2013counting}. Our experimental question is whether these conclusions—obtained when using the original data—can be faithfully reproduced when the control arm is replaced or augmented by synthetic data generated with state-of-the-art models.

\subsection{Control arm generation strategies}

More specifically, we consider two distinct scenarios, both of which correspond to pressing issues in clinical research:
\paragraph{Data sharing under privacy constraints}
An investigating team has access to the full clinical trial dataset, including $N^T$ treated and $N^C$ control patients. To enable external validation or meta-analysis, the control arm must be shared with another group. However, direct release of patient-level data is often prohibited by privacy regulations and data-use restrictions. In this setting, the task is to generate a synthetic control arm that closely reproduces the original distribution, preserving utility while ensuring confidentiality.
\paragraph{Control-arm augmentation}
In many trials, particularly for rare diseases, highly targeted therapies, or early-phase studies, the control group is much smaller than the treated group ($N^C \ll N^T$). This imbalance reduces statistical power and compromises treatment effect estimation. Here, synthetic data are used to augment the control arm, partially correcting the imbalance and improving the robustness of downstream analyses.

In our experiments, the generator is trained on a fraction $\upsilon \in \{1/3, 2/3, 1\}$ of the available control patients ($N_{\text{train}} = \upsilon N^C$). We then generate $N_{\text{gen}}$ synthetic control arms, each of size $N_{\text{sim}} = N^T$ (matching the treated group). When $\upsilon = 1$, this corresponds to the data-sharing scenario; when $\upsilon < 1$, it corresponds to augmentation, with effective control-arm expansion factors of $\times 3$ and $\times 1.5$. An alternative training strategy, where treated-arm data are also used, is described in Section~\ref{sec:extended_XP}.

\subsection{Competing algorithms}

To benchmark our HI-VAE-based generative models, we compared them to two state-of-the-art baselines proposed in \cite{norcliffe2023survivalgan} and available in the \texttt{synthcity} library \citep{qian2023synthcity2}: SurvivalGAN and SurvivalVAE. Both methods are based on a modular architecture where the generation of synthetic covariates is separated from the modeling of survival outcomes, allowing flexibility in the choice of generative model—either GAN- or VAE-based—and independent strategies for assigning event times and censoring. This separated two-stage modeling pipeline is a key difference with our proposed HI-VAE model, which jointly learns covariates and time-to-event outcomes within a unified latent framework. Other tabular data generators such as CTGAN, TVAE, or normalizing flows were compared to SurvivalGAN in \cite{norcliffe2023survivalgan} and found less effective for survival modeling, and were therefore excluded from our comparison. We refer to these two competing methods as \texttt{Surv-GAN} and \texttt{Surv-VAE} (see Appendix~\ref{apd:model_detail}).

\subsection{Training and hyperparameter tuning}

We train the models on a dataset $\mathcal{D}_{\text{train}}$ with the same structure as described in Section~\ref{sec:Method}. In our model, we optimize the parameters by maximizing the evidence lower bound (ELBO, Eq.~\ref{eq:ELBO}) in a batch learning setting with the Adam optimizer~\citep{kingma2014adam}. Moreover, we monitor training convergence using early stopping based on this ELBO to avoid overfitting. For competing methods with modular architectures, the parameters of each module are optimized in a sequential manner, see~\cite{norcliffe2023survivalgan} for more details. 

Hyperparameters - such as the learning rate, batch size, and latent state dimensions - are selected by minimizing the distance between the Kaplan–Meier survival curves estimated from the observed data and those derived from synthetic data. We perform this optimization using the Optuna framework~\citep{akiba2019optuna}, conducting up to 150 trials to ensure reliable selection of hyperparameters. We refer the reader to Appendix~\ref{apd:training} and \ref{subsec:influence_hyperopt} for full details of the hyperparameter selection process. 

\subsection{Metrics}
We evaluate the quality of the synthetic data along three dimensions—resemblance, utility, and privacy—using the metrics implemented in the \texttt{synthcity} package \citep{qian2023synthcity}. Our selection of metrics is guided by the scoping review of~\cite{kaabachi2025scoping}.

\paragraph{Data resemblance} 
We quantify the similarity between synthetic and original control arm data distributions using the Jensen-Shannon distance (\textit{J-S distance}). The J-S distance measures the dissimilarity between two probability distributions, with smaller values indicating closer resemblance (range from 0 to 1). 

\paragraph{Utility} 
We evaluate whether the synthetic data can replicate clinically relevant statistical patterns using the \textit{Survival curves distance}. The survival curves distance integrates the absolute difference between Kaplan–Meier survival curves estimated from the original and synthetic control arms, with smaller distances indicating better preservation of survival patterns. 

\paragraph{Privacy} 
Following \cite{steier2025synthetic}, we evaluate privacy preservation using the \textit{$K$-map score}. The $K$-map score measures the minimum group size of quasi-identifier combinations in the synthetic control arm, with higher values indicating lower re-identification risk and thus better privacy preservation. \\

\noindent Additional metrics, including the Kolmogorov–Smirnov test, an XGBoost classifier performance score, and the Nearest Neighbor Distance Ratio, are described in Appendix~\ref{apd:additional_metrics}.

\paragraph{Computation of expected type I errors and powers}
For the simulation experiments, we assess type I error and power by computing the proportion of rejected null hypotheses from the log-rank test (the standard nonparametric test for comparing survival curves between groups, \cite{fleming2013counting}) and comparing them with the theoretical (asymptotic) power given by the formula of \citet{schoenfeld1983sample}. Further details are provided in Appendix~\ref{apd:power_computation}.

\paragraph{Multiple test adjustment} For each Monte Carlo replication, we compute the p-values of the log-rank tests comparing survival curves between the original control and each generated control set, yielding $N_{\text{sim}}$ p-values in total. To account for multiple comparisons, we then apply the Benjamini–Hochberg (BH) procedure \cite{benjamini1995controlling} to control the false discovery rate (FDR) at a nominal level $\alpha=0.05$.

\subsection{Simulations}
\label{sec:simu}
To evaluate our method, we simulate $M=100$ independent Monte Carlo replications of a dataset obtained as follows. For each subject $i \in \{1,\dots,N\}$, covariates $x_i \in \mathbb{R}^d$ are sampled from a multivariate normal distribution with Toeplitz covariance. Continuous features remain unchanged, while binary ones are obtained by thresholding $x_i^j \leftarrow \mathbf{1}\{x_i^j > 0\}$. Treatments $e_i$ are drawn from $\mathcal{B}(1,0.5)$. Event times ($\tau_i$) and censoring times ($c_i$) follow Weibull distributions parameterized by covariates and a treatment coefficient $\beta \in \{0,0.2,\ldots,1.0\}$. Distributions are either fully independent or conditionally independent given covariates. By construction, $\beta=0$ indicates no risk difference between control and treated groups, with the effect size increasing in $\beta$. Further details are given in Appendix~\ref{apd:simu}.

\subsection{Clinical trial data}  
We further validate our approach on four phase III clinical trial datasets.
 \paragraph{NCT00119613} Placebo-controlled trial of darbepoetin alfa with platinum-based chemotherapy in extensive-stage small-cell lung cancer \citep{pirker2008safety}.
  \paragraph{NCT00113763} Trial of panitumumab plus chemotherapy versus chemotherapy alone in metastatic colorectal cancer \citep{van2007open}.
  \paragraph{NCT00339183} Trial of chemotherapy with or without panitumumab in metastatic or recurrent head and neck squamous cell carcinoma \citep{vermorken2013cisplatin}.
  \paragraph{ACTG 320} Trial showing triple-drug antiretroviral therapy reduced AIDS progression or death compared with dual therapy in HIV-infected patients \citep{hammer1997controlled}.

A summary of the main characteristics of the simulated and clinical datasets is provided in Table~\ref{tab:summary_dataset} and details on the distribution of the variables are provided in Appendix~\ref{apd:distribution_real_Data}. 

Although our empirical evaluation focuses on oncology and HIV datasets, the proposed framework is readily applicable to any clinical domain where the primary endpoint is of time-to-event type (e.g., overall survival, disease-free survival, progression-free survival), which represents a substantial share of phase III trials across therapeutic areas.

\begin{table}[h]
\caption{Dataset characteristics: number of control samples ($N^C$), treated samples ($N^T$), and number of static features by type: categorical ($d^{\text{cat}}$), positive-valued ($d^{\text{pos}}$), and continuous real-valued ($d^{\text{real}}$).}
\label{tab:summary_dataset}
\begin{center}
\begin{tabular}{lccccc} 
 \toprule
 \textbf{Dataset} & $N^C$ & $N^T$ & $d^{\text{cat}}$ & $d^{\text{pos}}$ & $d^{\text{real}}$\\ 
 \midrule
 Simulation & 300 & 300 & 6 & 0 & 6\\
 ACTG 320 & 577 & 574 & 5 & 2 & 1\\
 NCT00119613 & 236 & 235 & 7 & 0 & 2\\
 NCT00113763 & 475 & 470 & 10 & 0 & 2\\ 
 NCT00339183 & 260 & 260 & 5 & 0 & 1\\
 \bottomrule
\end{tabular}
\end{center}
\end{table}

\section{Results}\label{sec:results}

\subsection{Performance comparisons on classical metrics}\label{subsec:classical_metrics}
\begin{figure}[h!]
    \centering
    \includegraphics[width=1\columnwidth]{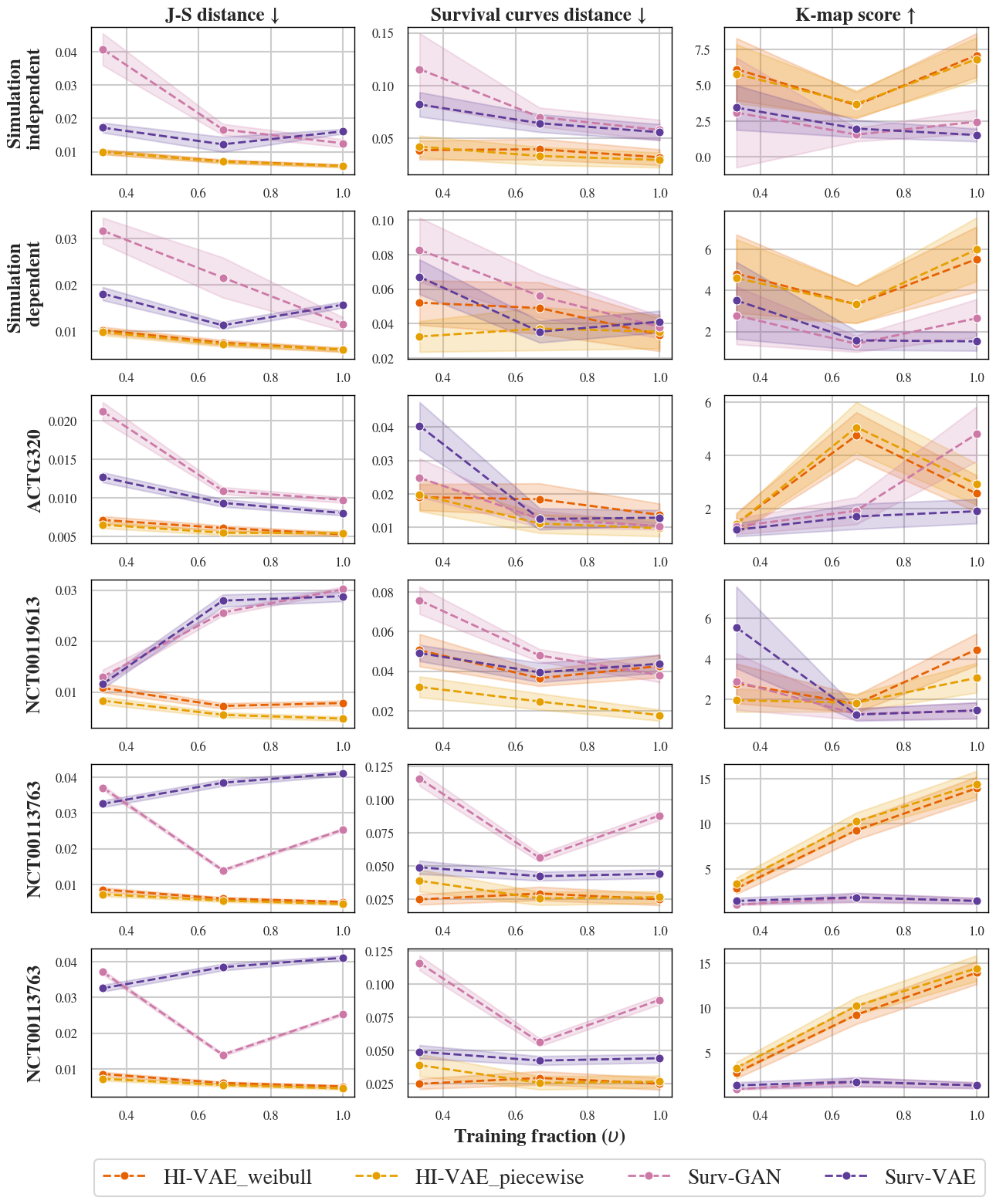}
   \caption{\footnotesize Performance comparison on simulated and real datasets, using J-S distance, survival curve distance, and $K$-map score. Arrows indicate directions of better performance. }
    \label{fig:perf_control}
\end{figure}

We first evaluated the performance of our algorithm in its two variants, \texttt{HI-VAE\_Weibull} and \texttt{HI-VAE\_piecewise}, against competing methods \texttt{Surv-GAN} and \texttt{Surv-VAE}, using the fidelity, utility, and privacy metrics described in Section~\ref{sec:experiments}. Figure~\ref{fig:perf_control} reports results on both simulated and real datasets. Our HI-VAE consistently achieved lower divergence from the original distributions and higher fidelity, particularly on survival outcomes, while maintaining competitive privacy scores. Supplementary results with additional metrics are provided in Appendix~\ref{apd:additional_metrics_perf}.

\subsection{Type I error and powers estimation in generated data} \label{subsec:results_clinical_research}
\begin{figure}[h!]
\centering
\includegraphics[width=1.\columnwidth]{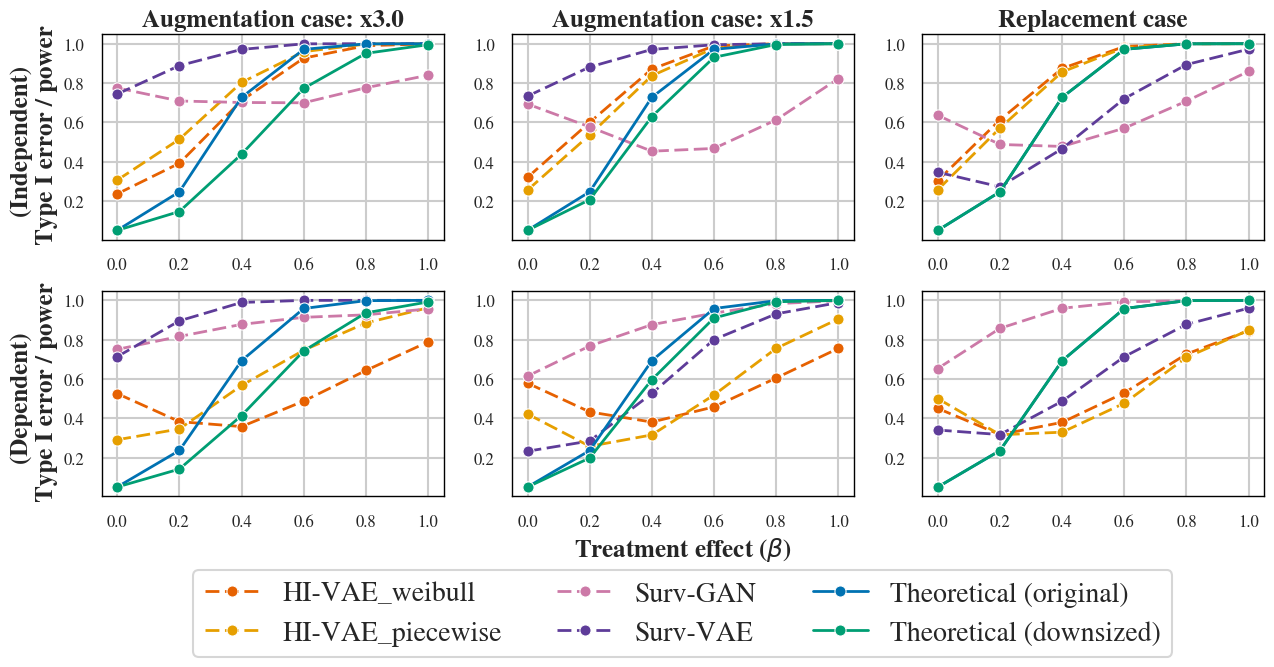}
\caption{\footnotesize Type I error and power estimation for \textbf{independent} case (\textbf{top}) and \textbf{dependent} case (\textbf{bottom}). Dashed lines: empirical power. Green: theoretical power with reduced control size. Blue: theoretical power with generated control size. }
\label{fig:simu_pvalue_control_dep_indep}
\end{figure}

We then assessed whether generative models reproduce the expected type I error ($\beta=0$) and statistical power ($\beta \in \{0.2,0.4,\dots,1.0\}$) predicted by Schoenfeld’s formula \citep{schoenfeld1983sample}. Figures~\ref{fig:simu_pvalue_control_dep_indep} summarize results for both independent and dependent cases over $M=100$ Monte Carlo replications.\\
Across all methods, type I error rates were consistently inflated, and power curves often deviated sharply from theoretical expectations, highlighting that good fidelity scores or usual utility metrics do not guarantee valid downstream inference.

\subsection{Post-generation selection} \label{subsec:post_generation_selection}

\begin{figure}[h!]
    \centering
    \includegraphics[width=1.0\columnwidth]{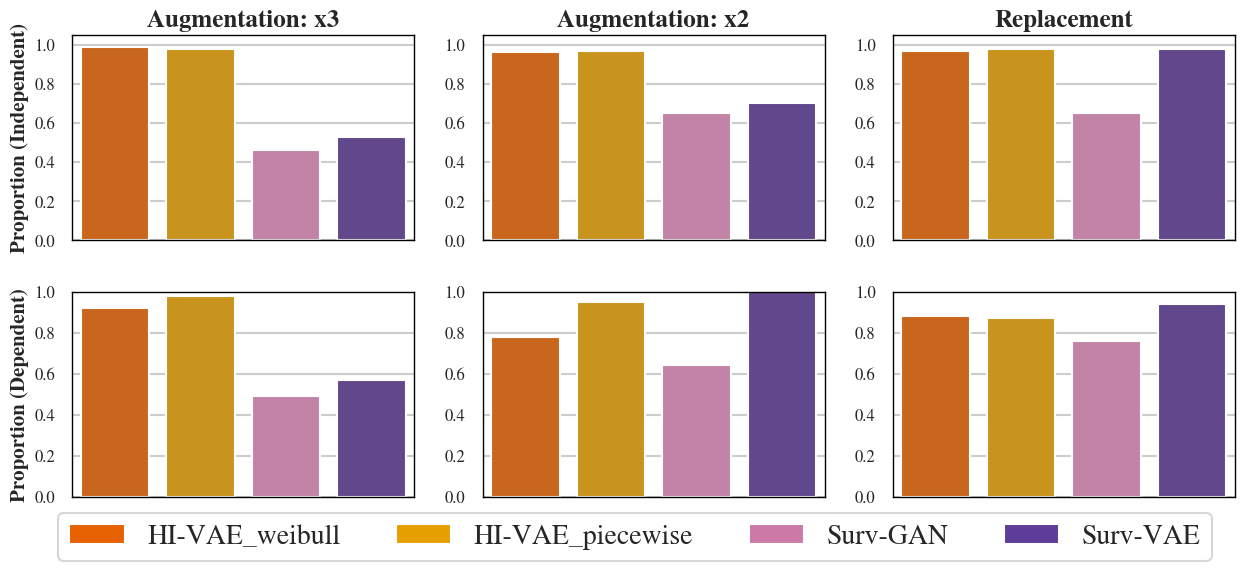}
 \caption{\footnotesize Proportion of Monte Carlo replications with at least one generated dataset not rejected by the adjusted log-rank test (at the 5\% level) against the original controls.}
    \label{fig:simu_compare_H0_v2} 
\end{figure}

We next investigated the reasons for the poor statistical performance of the generative algorithms. Figure~\ref{fig:simu_compare_H0_v2} shows the proportion of Monte Carlo replications in which at least one of the $N_{\text{gen}} = 200$ generated datasets for which the multiple-testing–adjusted log-rank test of equal survival between original and generated controls was not rejected. Our method achieved this in at least 80\% of replications across all experimental settings—and often considerably more—making it substantially more reliable and robust than competing approaches.


We then retained, for each case, the best generated control dataset from among the $N_{\text{gen}} = 200$ candidates. “Best” was defined as the dataset yielding the highest $p$-value in a log-rank test comparing survival distributions between the original training and generated controls. In the simulation study, this procedure resulted in one selected dataset per Monte Carlo replication. Further details on the selection procedure and on the computation of type I error rates and power are provided in Appendix~\ref{apd:power_computation}.

Figure~\ref{fig:simu_pvalue_control_best_merge_indep_dep} reports the type I error rates and powers for both independent and dependent settings. In the independent case, the datasets generated by our algorithm closely reproduce the target theoretical power for a control size of $N^C = 300$ (blue line), even when trained on as few as $N_\text{train} = 100$ or $200$ original controls. This indicates a genuine augmentation effect, particularly in the $\times 3$ augmentation scenario. By contrast, this effect is attenuated or inexistent in the dependent setting. Type I error rates are, however, somewhat inflated in augmentation settings, with inflation increasing with the augmentation factor.


\begin{figure}[h!]
\centering
\includegraphics[width=1.0\columnwidth]{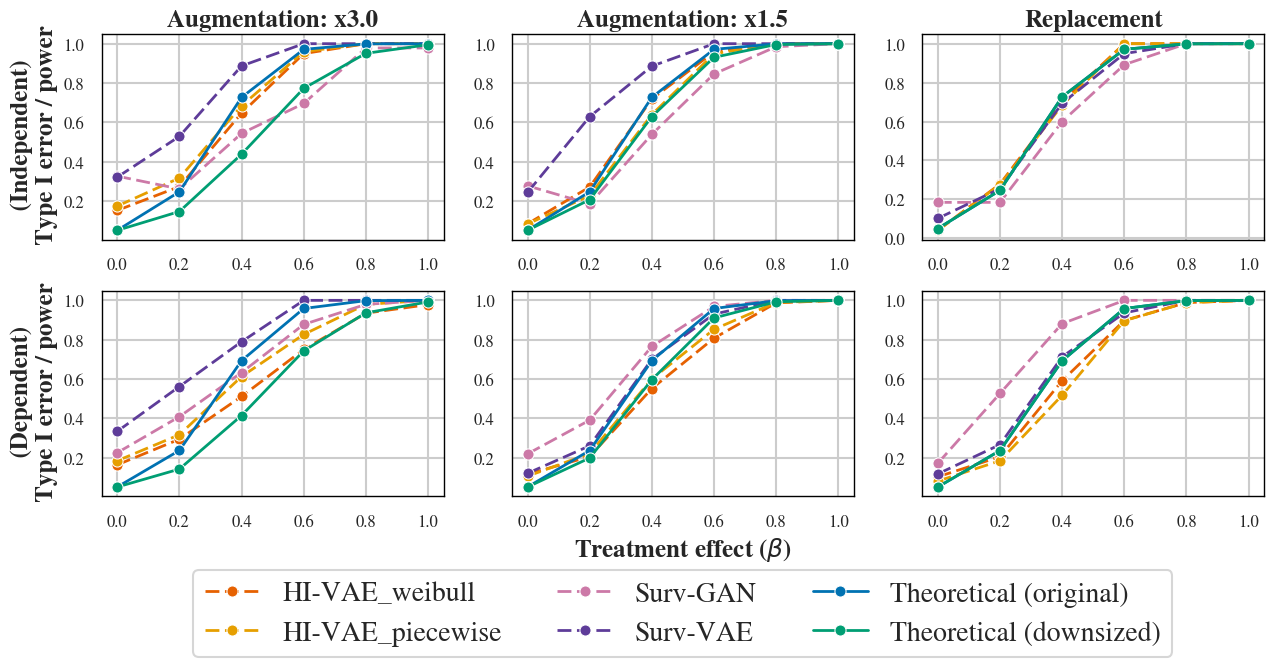}
\caption{\footnotesize Type I error and power estimation after post-generation selection for \textbf{independent} case (\textbf{top}) and \textbf{dependent} case (\textbf{bottom}). Dashed lines: empirical power. Green: theoretical power with reduced control size. Blue: theoretical power with generated control size.}
\label{fig:simu_pvalue_control_best_merge_indep_dep}
\end{figure}

Conversely, the \texttt{Surv-GAN} algorithm generally fails to reproduce the expected type I error rates or powers, possibly due to the limited proportion of generated control survival data resembling the originals. The \texttt{Surv-VAE} algorithm attains the target theoretical power in three of the six experiments, most notably in the replacement scenario, but falls short in the others and does not adequately control type I error in most cases. Its heterogeneous performance across experiments may partly reflect variability in how closely the generated control survival data approximate the originals. These observations are consistent with Figure~\ref{fig:simu_compare_H0_v2}.

Importantly, other performance metrics are not degraded by the selection procedure (see Figure~\ref{fig:compare_simu_perf_control} in Appendix~\ref{apd:metrics_best}).

Finally, reducing the candidate pool (e.g., to 50 or 100 datasets instead of 200) does not alter these conclusions (see Figure~\ref{fig:simu_perf_control_best_vs_all} in Appendix~\ref{apd:metrics_best}). However, selecting the top 20\% of generated datasets leads to poor performance for all methods (Figure~\ref{fig:simu_pvalue_control_best_40_indep_dep} in Appendix~\ref{apd:metrics_best}).


\subsection{Extended experiments}
\label{sec:extended_XP}

\paragraph{Risk-model discrimination and calibration}
To compare the discrimination and calibration of the risk model trained on synthetic versus real control data, we performed an additional experiment in which the Cox proportional hazards model \citep{Cox1972JRSS} was trained separately on the two datasets. The predictive performance was then evaluated on an independent real test set using the C-index~\citep{harrell1996tutorial} and the integrated Brier score~\citep{graf1999assessment}. As shown in Appendix~\ref{subsec:discrimination/calibration} (Figures~\ref{fig:real-Cindex} - \ref{fig:real-BS}), the results indicate that our method produces synthetic-trained models whose performance closely matches that of real controls in both discrimination and calibration, whereas Surv-GAN and Surv-VAE achieve good discrimination but exhibit poorer calibration.



\paragraph{Impact of hyperparameter optimization}


We conducted additional experiments to assess the impact of both the hyperparameter search strategy and the random seed within the Optuna framework. As detailed in Appendix~\ref{subsec:influence_hyperopt} (Figures~\ref{fig:simu_pvalue_comp_method_hyperopt}–\ref{fig:simu_perf_comp_seed_hyperopt}), we compared three search methods (1–3) for two models (\texttt{HI-VAE\_piecewise} and \texttt{Surv-VAE}) in the independent setting, training only on the control arm.

Overall, strategies that evaluated performance on generated synthetic datasets of full control size (methods 1 and 3) proved more robust than relying solely on validation-set evaluation, which can be limited in size (method 2). With these approaches, the relative performance of our model versus \texttt{Surv-VAE} remained stable across random seeds. However, seed choice still influenced the absolute performance within each algorithm, highlighting the sensitivity to hyperparameter selection. In our experiments, we fixed the number of Optuna trials to 150, but in practice, a more extensive search would likely be needed to ensure optimal results.

\paragraph{Training setup: control vs. control + treated}

In our initial experiments, the generator was trained exclusively on the available controls—or on a subset thereof in the augmentation setting. We then explored an alternative strategy in which treated-arm data were also incorporated during training, as illustrated in Figure~\ref{fig:generation_testing} of Appendix~\ref{apd:generation}. The rationale was that in clinical trial contexts, where datasets are often small (e.g., $\upsilon = 1/3$ with 100 controls and 300 treated), leveraging treated data might yield more robust models.

\begin{figure}[h!]
\centering
\includegraphics[width=1.0\columnwidth]{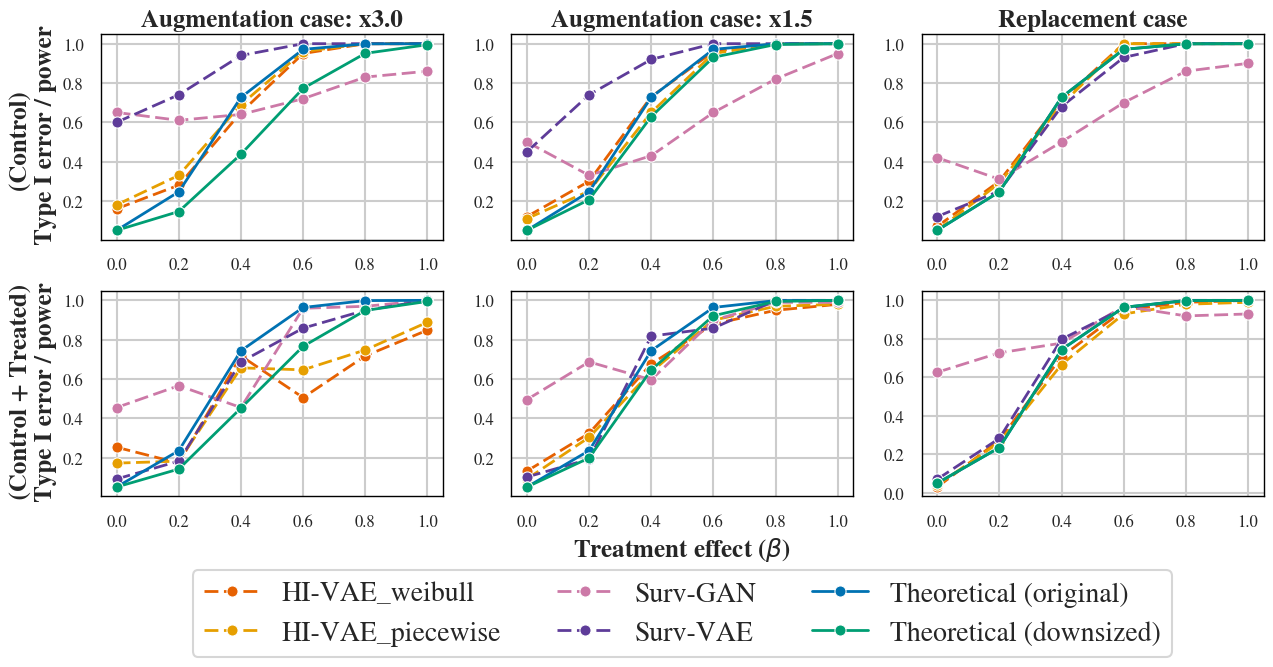}
\caption{\footnotesize Type I error and power estimation after post-generation selection for independent case under two training strategies: using only available control samples (\textbf{top}) and using both control and treated arms (\textbf{bottom}). Dashed lines: empirical power. Green: theoretical power with reduced control size. Blue: theoretical power with generated control size.}
\label{fig:simu_pvalue_control_best_merge_control_full}
\end{figure}

However, results on both simulated and real datasets did not clearly support this intuition (see Figure~\ref{fig:simu_pvalue_control_best_merge_control_full} and Appendix~\ref{subsec:control_vs_full}). On simulated data, resemblance and utility metrics were generally superior when training on controls only, whereas privacy metrics improved when both controls and treated were used—consistent with the expected utility–privacy trade-off. Among the models, \texttt{Surv-VAE} was the only one to show a modest benefit from including treated data; the others did not. On real datasets, all models performed better on average when trained solely on controls, particularly with respect to data resemblance metrics.


\paragraph{Privacy considerations: prior sampling and differential privacy}

Our privacy analysis showed that posterior-based sampling falls short of the thresholds required for effective data sharing under current standards such as EMA Policy 0070~\citep{ema_policy_0070}, which recommends $K$-map values above 11 for public release. In our setting, posterior sampling yielded $K$-map values of about 4–6 and NNDR values around 0.2–0.4. These levels may still be acceptable under controlled-access regimes (e.g., data use agreements) or for augmentation scenarios, but they remain insufficient for regulatory-grade data sharing. To address this gap, we explored two complementary strategies.

\begin{figure}[h!] \centering \includegraphics[width=1.0\columnwidth]{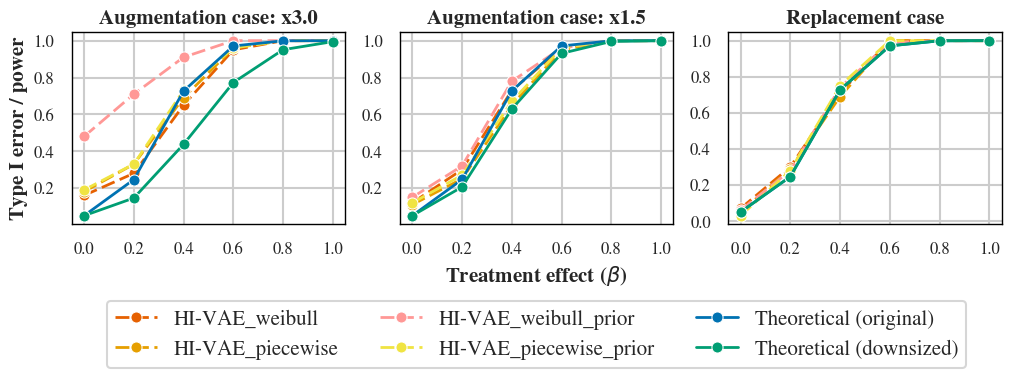} \caption{\footnotesize Type I error and power estimation after post-generation selection for independent case. Dashed lines: empirical power. Green: theoretical power with reduced control size. Blue: theoretical power with generated control size.} \label{fig:simu_pvalue_control_best_merge_prior_posterior} \end{figure}

The first strategy was to replace posterior sampling with prior-based sampling. Unlike posterior sampling, which draws latent variables conditional on the observed data, prior sampling generates them directly from the model priors, producing fully synthetic samples that are in principle less dependent on the original dataset and therefore potentially more privacy-preserving. In practice, however, we found only minor differences between the two approaches. On independent simulated datasets, posterior sampling performed slightly better overall (Figure~\ref{fig:simu_pvalue_control_best_merge_prior_posterior}, Table~\ref{tab:privacy_metrics_prior_posterior}). On real datasets, results were more mixed: prior sampling improved some $K$-map scores but not NNDR values. Overall, we did not observe consistent privacy benefits from switching to prior-based sampling.

\begin{table}[!h] \scriptsize \centering \setlength{\tabcolsep}{3.2pt} 
\renewcommand{\arraystretch}{1.1} 
\begin{tabular}{c|c|c|cc} \hline \textbf{Dataset} & \texttt{HI-VAE} & \textbf{Metric} & \textbf{Post.} & \textbf{Prior} \\ \hline \multirow{4}{*}{\makecell{Simulation \\ (independent)}} & \multirow{2}{*}{\texttt{\_piecewise}} & $K$-map & $\mathbf{5.44 \pm 3.41}$ & $4.97 \pm 3.17$ \\ & & NNDR & $\mathbf{0.42 \pm 0.07}$ & $0.41 \pm 0.07$ \\ & \multirow{2}{*}{\texttt{\_weibull}} & $K$-map & $\mathbf{5.61 \pm 3.57}$ & $5.03 \pm 3.62$ \\ & & NNDR & $\mathbf{0.42\pm0.07}$ & $0.41 \pm 0.06$ \\ \hline \multirow{4}{*}{NCT00119613} & \multirow{2}{*}{\texttt{\_piecewise}} & $K$-map & $3.07 \pm 1.45$ & $\mathbf{3.49 \pm 1.70}$ \\ & & NNDR & $\mathbf{0.18 \pm 0.04}$ & $ 0.17 \pm 0.04$ \\ & \multirow{2}{*}{\texttt{\_weibull}} & $K$-map & $ 4.47 \pm 1.57$ & $\mathbf{4.86 \pm 1.53}$ \\ & & NNDR & $0.21 \pm 0.05$ & $\mathbf{0.22 \pm 0.04}$ \\ \hline 
\end{tabular} \caption{\footnotesize Comparison of $K$-map ($\uparrow$) and NNDR ($\uparrow$) scores for posterior vs. prior sampling in the replacement setting ($\upsilon = 1$), evaluated on the independent simulated dataset and the NCT00119613 dataset.} \label{tab:privacy_metrics_prior_posterior} \end{table}

The second strategy was a preliminary application of differential privacy using the \texttt{Opacus} framework~\citep{opacus}. We trained the HI-VAE models with per-step noise injection (\texttt{noise\_multiplier = 2.0}). Under this configuration, privacy metrics did not improve substantially compared to the non-private models (Appendix~\ref{subsec:differential_privacy}). Stronger protection would likely require more aggressive parameter settings, which in turn would degrade utility.

In summary, while synthetic patients generated with posterior or prior sampling are not direct replicas of real ones, neither approach provided sufficient privacy guarantees for open data release. Differential privacy offered no improvement under moderate noise levels, underscoring the difficulty of balancing utility and privacy in this setting.

\section{Conclusion}
We introduced a VAE-based framework for generating synthetic control arms with time-to-event outcomes. Across synthetic and real clinical trial datasets, and under both data-sharing and data-augmentation scenarios, our method outperformed baselines in terms of fidelity, utility, and privacy. However, a key limitation emerged: despite strong performance on classical metrics, all models—including ours—yielded miscalibrated survival analyses, with inflated type I error and biased power. Post-generation selection improved calibration and restored power in most settings, though type I error remained partially elevated.

These findings highlight the importance of evaluating generative models for health not only on fidelity and privacy, but also on their downstream statistical calibration. A model that looks strong by standard ML metrics may still fail when applied to clinical inference.

Our privacy analysis also underscored important limitations. Posterior sampling yielded $K$-map values consistently below the EMA Policy 0070~\citep{ema_policy_0070} benchmark ($\geq 11$), confirming that current configurations remain insufficient for public data release. Nevertheless, such values may still be acceptable under controlled-access regimes (e.g., data use agreements) or in augmentation scenarios. Complementary metrics (NNDR, detection tests) indicated only partial protection against re-identification. A preliminary attempt to integrate differential privacy (Opacus) did not yield substantial gains, underscoring the sensitivity of privacy–utility trade-offs to parameter choices.

Future work should explore stronger privacy-preserving techniques—such as calibrated $\varepsilon$–$\delta$ differential privacy, diffusion- or transformer-based generators with privacy-aware objectives—and the development of calibration-aware training strategies. However, adapting these models to survival settings requires further methodological development. More broadly, bridging generative modeling with domain-specific notions of validity, such as error control in survival analysis, will be essential to ensure reliability in real-world applications.\\
In summary, we provide the first systematic evaluation of type I error and power in generative survival models. Our results demonstrate both the promise of VAEs for survival data generation and the need for methodological advances before such models can be safely applied in clinical research.

\section*{Acknowledgments \& Fundings}
For this work, AG and VTN have benefited from the support of the National Agency for Research under the France 2030 program with the reference ANR-22-PESN-0016. PC, LD and AG received funding from the MEDITWIN Bpifrance i-Demo project as part of the France 2030 program. AG has also benefited from the Institut National du Cancer. All authors thank Armelle Arnoux and Sandrine Katsahian, who helped design the experimental settings.

\bibliography{jmlr-sample}

\onecolumn
\appendix
\section{Supplementary materials for Section~\ref{sec:Method}}
\subsection{Algorithmic and Implementation Details}\label{apd:model_detail}

\paragraph{Details on the feature-specific conditional densities}

\begin{itemize}
    \item Continuous real-valued variables (Normal): $h_j(\mathbf{y}_i, \mathbf{s}_i) = \big(\mu_j(\mathbf{y}_i, \mathbf{s}_i), \sigma_j^2(\mathbf{y}_i, \mathbf{s}_i)\big)$, 
    \item Positive real-valued variables (Log-Normal): $h_j(\mathbf{y}_i, \mathbf{s}_i) = \big(\mu_j(\mathbf{y}_i, \mathbf{s}_i), \sigma_j^2(\mathbf{y}_i, \mathbf{s}_i)\big)$, 
    \item Count variables (Poisson):  $h_j(\mathbf{y}_i, \mathbf{s}_i) = \lambda_j (\mathbf{y}_i, \mathbf{s}_i)$,
    \item Categorical variables (Multinomial-logit): $h_j(\mathbf{y}_i, \mathbf{s}_i) = \big(h_{j0}(\mathbf{y}_i, \mathbf{s}_i), \hdots, h_{j(R-1)}(\mathbf{y}_i, \mathbf{s}_i)\big)$.
\end{itemize}

\paragraph{Details on the parameterization of the time-to-event density}

We consider two parameterizations of the density $p$ in our implementation. In the \texttt{HI-VAE\_Weibull} variant, the density $p$ is modeled using a Weibull distribution. For each subject, the neural network $\eta(\mathbf{y}_i, \mathbf{s}_i)$ outputs the scale and shape parameters, denoted as $\text{sc}(\mathbf{y}_i, \mathbf{s}_i)$ and $\text{sh}(\mathbf{y}_i, \mathbf{s}_i)$, respectively. The survival function $\bar P$ then writes $$\bar P(t) = \exp\left(-\left(\tfrac{t}{\text{sc}}\right)^{\text{sh}}\right).$$
In the \texttt{HI-VAE\_piecewise} variant, we follow~\cite{friedman1982piecewise,lee2018deepHit,kvamme2019continuous,cottin2022idnetwork} and discretize the time axis into $K$ disjoint intervals $i_1,\ldots,i_K$, with the right endpoint of $i_K$ chosen larger than the maximal observed time. A neural network $\eta(\mathbf{y}_i,\mathbf{s}_i)$ with a softmax output layer provides interval probabilities $(p_1,\ldots,p_K)$ (with $\sum_{k=1}^K p_k = 1$). The survival function $\bar P$ is then given by
\[
1- \bar P(t) = \sum_{k=1}^{i(t)-1} p_k + \frac{(t - i_{i(t)-1})}{|i_{i(t)}|}p_{i(t)},
\]
where $|i_k|$ denotes the length of the interval $i_k$ and $i(t)$ the index of the interval containing $t$.

\paragraph{More details on survivalGAN and survivalVAE}

SurvivalGAN first encodes covariates via Gaussian mixture and one-hot encodings, then constructs a conditional GAN (based on ADS-GAN \citep{Yoon2020adsgan}) to generate synthetic covariates guided by a compact condition vector. A survival function (parameterized by DeepHit \citep{lee2018deepHit}) estimates survival probabilities, followed by a time regressor that outputs event or censoring times. 
As a result, synthetic samples are generated in sequential stages: sampling latent noise, generating covariates, estimating survival curves, and regressing event times.
SurvivalVAE adopts the same pipeline framework, replacing the GAN module with a Tabular Variational Autoencoder \citep{Xu2019} to generate covariates.

\subsection{Details on the control arm generation process}
\label{apd:generation}


We provide in Figure~\ref{fig:generation_testing} additional details on how synthetic control arms are generated, illustrating both training on the control arm only and training on the combined control and treated arms (corresponding to additional experiments in Section~\ref{sec:extended_XP}).

\subsection{Hyperparameter optimization}
\label{apd:training}
We list the hyperparameter search space for each algorithm.

\begin{itemize}
    \item \texttt{HI-VAE}. Learning rate: \{$2\mathrm{e}^{-2}$, $\mathrm{e}^{-3}$, $\mathrm{e}^{-4}$\}; 
    batch size: $\{0.25, 0.4, 0.6, 0.75\} \times N_{\text{train}} \cup \{100\}$; 
    latent dimensions: $z \in [10,200]$ (step 10), $y \in [10,200]$ (step 5), $s \in [10,200]$ (step 10); 
    number of survival layers (for \texttt{HI-VAE\_piecewise} only): $\{1,2\}$; 
    number of piecewise intervals (for \texttt{HI-VAE\_piecewise} only): $\{5,10,15,20\}$.

    \item \texttt{Surv-VAE}. (from \texttt{synthcity} \cite{qian2023synthcity}) Number of max epochs $\{100,200,300,400,500\}$; 
    learning rate: $\{\mathrm{e}^{-3},2\mathrm{e}^{-4},\mathrm{e}^{-4}\}$;
    weight decay: $\{\mathrm{e}^{-3},\mathrm{e}^{-4}\}$;
    batch size: $\{64,128,256,512\}$;  
    embedding units: $[50,500]$ (step 50); 
    encoder/decoder: hidden layers $\in[1,5]$, hidden units $\in [50,500]$ (step 50);
    nonlinearities: $\{\texttt{relu},\texttt{leaky\_relu},\texttt{tanh},\texttt{elu}\}$;  
    dropout $\in[0,0.2]$.

    \item \texttt{Surv-GAN}. (from \texttt{synthcity} \cite{qian2023synthcity}) 
    Learning rate: $\{\mathrm{e}^{-3},2\mathrm{e}^{-4},\mathrm{e}^{-4}\}$; weight decay: $\{\mathrm{e}^{-3},\mathrm{e}^{-4}\}$; encoder clusters $\in[2,20]$;
    Generator: hidden layers $\in[1,4]$,
    hidden units $\{50,100,150\}$,
    nonlinearities $\{\texttt{relu},\texttt{leaky\_relu},\texttt{tanh},\texttt{elu}\}$,
    dropout $\in[0,0.2]$; 
    Discriminator: hidden layers $\in[1,4]$, 
    hidden units $\{50,100,150\}$, same nonlinearities, dropout $\in[0,0.2]$.
\end{itemize}


\subsection{Additional evaluation metrics} 
\label{apd:additional_metrics}

\paragraph{Data resemblance} 
The Kolmogorov-Smirnov test (\textit{KS test}) compares the empirical cumulative distribution functions of synthetic and original features, producing a score between 0 and 1, where 1 indicates identical distributions.

\paragraph{Utility} 
We also compute a performance score by training an XGBoost classifier (\textit{Detection XGB}) to distinguish between original and synthetic samples. A result of 0 means the two datasets are indistinguishable (best), and 1 means they are completely distinguishable (worst for privacy).

\paragraph{Privacy} 
We additionally compute the Nearest Neighbor Distance Ratio (\textit{NNDR}), which compares, for each real sample, the distance to its nearest synthetic neighbor relative to the distance to its nearest real neighbor. 
A score of 0 indicates exact reproduction of real samples in the synthetic dataset (privacy leakage), while a score of 1 indicates that all synthetic samples are far from any real data (stronger privacy).

\subsection{Type I error and power computations}
\label{apd:power_computation}

\paragraph{Monte Carlo experiments and estimations of type I error and powers}

In our Monte Carlo experiments, for each value (size effect of the treatment $e_i$) $\beta$ in $\{0,0.2,0.4,0.6,0.8,1.0\}$, we generated $M=100$ control and treated arms of resp. sizes $N^C$ and $N^T$. We obtained $M$ initial p-values for the log-rank tests performed on each replication: $\text{pv}\big(\mathcal  D_\text{control,1} , \mathcal  D_{\text{treated},1},\beta \big)$ to $\text{pv}\big(\mathcal  D_\text{control,M} , \mathcal  D_{\text{treated},M},\beta \big)$. We then computed an approximate power
\begin{align}
    \text{power}_{\text{inital}}(\beta) = \frac1M \sum_{m=1}^M  \mathbf{1}\{\text{pv}\big(\mathcal  D_\text{control,m} , \mathcal  D_{\text{treated},m},\beta \big) < 0.05 \},
\end{align}
notice that, for $\beta=0$, it corresponds to
the approximate type I error.

In a second time, for each replication $m$, we generated $N_{\text{gen}}$ synthetic control arms, $\mathcal D^1_\text{gen,m}$ to $\mathcal D^{N_{\text{gen}}}_\text{gen,m}$, from $\mathcal  D_{\text{control},m}$ the initial control arm (and the $\mathcal  D_{\text{treated},m}$ treated arm) as depicted in Figure~\ref{fig:generation_testing}. For each of them, we computed the log-rank test p-value
$\text{pv}\big(\mathcal  D^n_\text{gen,m} , \mathcal  D_{\text{treated},m},\beta \big)$. This is summarized for one Monte Carlo experiment in Figure~\ref{fig:generation_testing}. We then defined the approximate power (resp. type I level) reached by the synthetic control arms as
\[
    \text{power}_{\text{gen}}(\beta) = \frac{1}{M N_{\text{gen}}} \sum_{m=1}^M  \sum_{n=1}^{N_{\text{gen}}}\mathbf{1}\{\text{pv}\big(\mathcal  D^n_\text{gen,m} , \mathcal  D_{\text{treated},m},\beta \big) < 0.05 \},
\]
these values are reported in Figure~\ref{fig:simu_pvalue_control_dep_indep}. 

Finally, we selected only the best generated control dataset in each case (among $N_{\text{gen}} = 200$ candidates) by comparing the $p$-values in a log-rank test comparing survival distributions between the original training and generated controls.
\[ n_{\text{best},m} = \text{argmin}_{n = 1, \ldots,N_{\text{gen}}} \; \text{pv}\big(\mathcal  D^n_\text{gen,m} , \mathcal  D_{\text{control},m}\big). \]
We then compute the approximate type I error and power reached by these selected datasets by appliying the following formula
\[     \text{power}_{\text{gen,best}}(\beta) = \frac{1}{M } \sum_{m=1}^M \mathbf{1}\{\text{pv}\big(\mathcal  D^{n_{\text{best},m} }_\text{gen,m} , \mathcal  D_{\text{treated},m},\beta \big) < 0.05 \}. \]
\begin{center}
\begin{figure}[!h]
\begin{tikzpicture}[
  >=Latex,
  font=\scriptsize,
  scale=0.95, every node/.style={transform shape},
  node distance=0.6cm and 1.0cm,
  box/.style={
    draw, rounded corners=1.2mm, minimum width=2.6cm, minimum height=0.6cm,
    align=center, fill=#1!20
  },
  diamondblock/.style={
    draw, diamond, aspect=2, align=center, minimum width=1.4cm,
    minimum height=0.8cm, fill=gray!15
  },
  stepbox/.style={
    draw, rounded corners=1.2mm, minimum width=2.6cm, minimum height=0.6cm,
    align=center, fill=orange!20
  },
    test/.style={
      draw, rounded corners=1.2mm, minimum width=2.6cm, minimum height=0.7cm,
      align=center, fill=testViolet!15
    }
]

\colorlet{origGreen}{green!70}
\colorlet{synGreen}{green!40}
\colorlet{treatRed}{red!70}
\colorlet{testViolet}{violet!70}

\node[box=origGreen] (origCtrl) {Original control arm\\ $\mathcal D_\text{control}$ ($N^C$)};
\node[box=testViolet, right=3.5cm of origCtrl, minimum width=2.2cm] (initial) {Initial log-rank test\\$\text{pv}\big(\mathcal  D_\text{control} , \mathcal  D_{\text{treated}}\big)$};
\node[box=treatRed, right=2.5cm of initial] (origTreat) {Original treated arm\\ $\mathcal D_\text{treated}$ ($N^T$)};

\draw[->, line width=0.6pt] (origCtrl.east) -- (initial.west);
\draw[<-, line width=0.6pt] (initial.east) -- (origTreat.west);

\node[stepbox, below=0.9cm of origCtrl] (downsizing) {Downsizing to $\mathcal D^{\text{train}}_{\text{control}}$ \\ ($N_\text{train} = \upsilon \times N^C$)};
\draw[->, line width=0.6pt] (origCtrl.south) -- (downsizing.north);

\node[diamondblock, below=0.9cm of downsizing] (train) {Training\\
Hyperparameter tuning\\
Generation};
\draw[->, line width=0.6pt] (downsizing.south) -- (train.north);

\draw[dashed, ->, line width=0.6pt] 
  (origTreat.south) |- (train.north)
  node[pos=0.55, above left] {(when traited arm is used for training)};
\node[box=synGreen, below=1.5cm of train, xshift=1.2cm] (syn1) {Synthetic control arm\\ $\mathcal D_\text{gen}^1$ $(N_{\text{sim}})$    };

\node[box=synGreen, below=0.7cm of syn1] (syn2) {Synthetic control arm\\ $\mathcal D_\text{gen}^2$ $(N_{\text{sim}})$  };

\node at ($(syn2.south) + (0,-0.8)$) {$\vdots$};

\node[box=synGreen, below=1.9cm of syn2] (syn3) {Synthetic control arm\\ $\mathcal D_\text{gen}^{N_{\text{gen}}}$ $(N_{\text{sim}})$};

\coordinate (Lbus) at ($(syn1.west)+(-0.9,0)$);
\draw[line width=0.6pt] (Lbus |- syn1.west) -- (Lbus |- syn3.west);
\coordinate (Lmid) at ($(Lbus |- syn1.west)!0.5!(Lbus |- syn3.west)$);
\draw[-, line width=0.6pt] (train.south) -| (Lmid);
\foreach \s in {syn1,syn2,syn3}{
  \draw[-, line width=0.6pt] (Lbus |- \s.west) -- (\s.west);
}

\node[test, below=6.85cm of initial] (test1) {Log-rank test\\$\text{pv}\big(\mathcal  D^1_\text{gen} , \mathcal  D_{\text{treated}}\big)$};
\node[test, below=0.69cm of test1] (test2) {Log-rank test\\$\text{pv}\big(\mathcal  D^2_\text{gen} , \mathcal  D_{\text{treated}}\big)$};
\node[test, below=1.9cm of test2] (test3) {Log-rank test\\$\text{pv}\big(\mathcal  D^{N_{\text{gen}}}_\text{gen} , \mathcal  D_{\text{treated}}\big)$};

\draw[->, line width=0.6pt] (syn1.east) -- (test1.west);
\draw[->, line width=0.6pt] (syn2.east) -- (test2.west);
\draw[->, line width=0.6pt] (syn3.east) -- (test3.west);

\coordinate (Rbus) at ($(origTreat.west)+(-0.7,0)$);
\draw[line width=0.6pt] (Rbus |- test1.east) -- (Rbus |- test3.east);
\foreach \t in {test1,test2,test3}{
  \draw[<-, line width=0.6pt] (\t.east) -- (Rbus |- \t.east);
}
\coordinate (Rmid) at ($(Rbus |- test1.east)!0.5!(Rbus |- test3.east)$);
\draw[-, line width=0.6pt] (origTreat.south) |- (Rmid);
\end{tikzpicture}
\caption{Workflow for generating synthetic control arms and evaluating treatment effects. Original control data are downsized and used to train the generator (optionally including treated data). Multiple synthetic control replicates are then produced and compared with the treated arm using log-rank tests.}
\label{fig:generation_testing}
\end{figure}
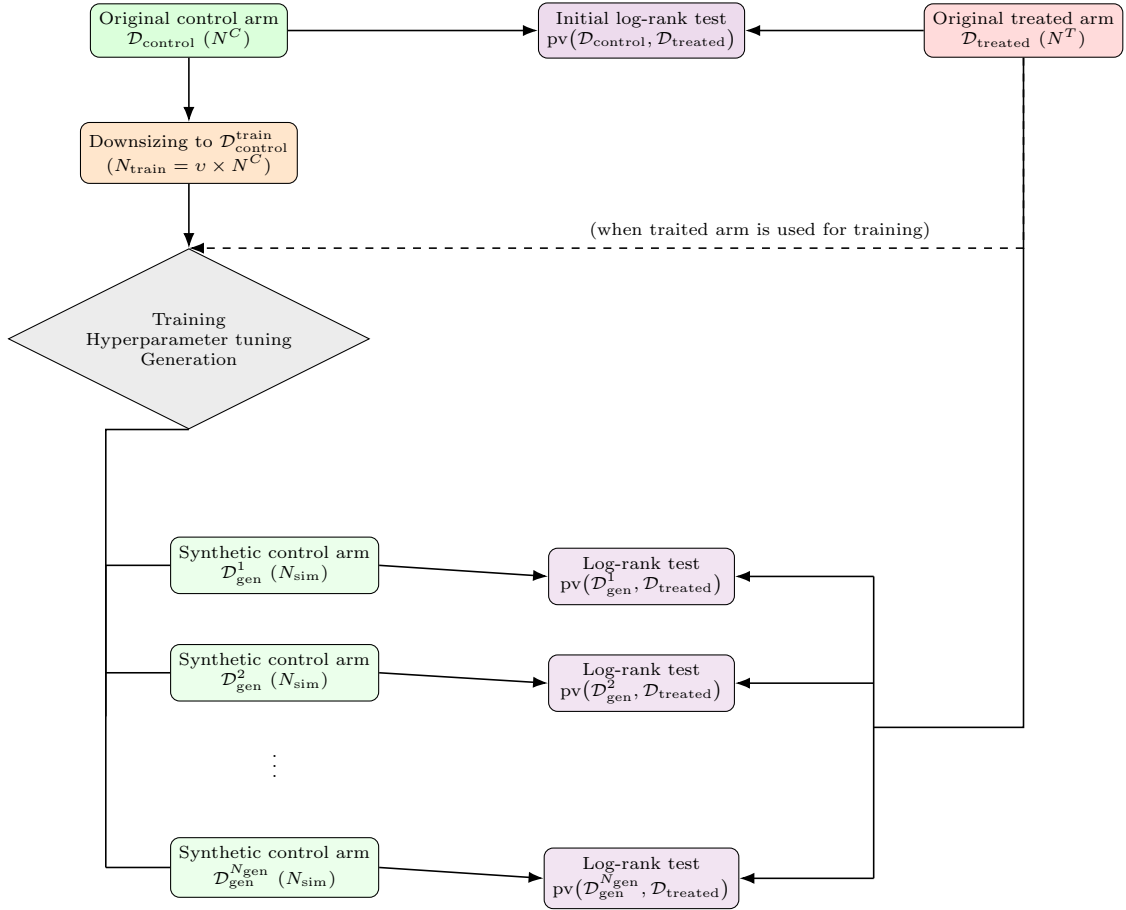
\end{center}

\paragraph{Theoretical power formula from~\cite{schoenfeld1983sample}}

Wherever reported, the theoretical power has been computed the following way, following the implementation of the function \texttt{cpower}
 of the  \texttt{Hmisc} R package \citep{harrell2019package}.

 \begin{align*}
 \text{power}(\tilde \beta) = 1 - \Big(\Phi\big(\Phi^{-1}(\alpha/2) - \frac{|\tilde \beta | }{ \sigma} \big) - \Phi\big(-\Phi^{-1}(\alpha/2) - \frac{|\tilde \beta |}{\sigma}\big)\Big)
 \end{align*}
 where
 \begin{itemize}
 \item $\Phi$ is the cumulative distribution function of the standard Normal,
     \item $\sigma = \sqrt{\frac{1}{\text{ns}_T} + \frac{1}{\text{ns}_C} }$ with $ \text{ns}_T , \text{ns}_C $ are the (average) number of survivors in the treated and control groups
     \item $\tilde \beta$ is the univariate equivalent of the treatment coefficient $\beta$ (computed via Monte Carlo experiments in each simulation setting).
 \end{itemize}

\subsection{Details on the simulation settings}\label{apd:simu}

The static covariates $x_i \in \mathbb{R}^d$ from a multivariate normal distribution with Toeplitz covariance $\Sigma \in \mathbb{R}^{d \times d}$:
\begin{align*} x_i \sim \mathcal{N}(0, \Sigma), \quad \Sigma_{jk} = \rho^{|j-k|}, \quad j,k = 1,\dots,d. \end{align*} 

\noindent Event ($\tau_i$) and censoring ($c_i$) times are simulated as
\begin{align*}
    \tau_i &= (-\log(1 - u_i) / \exp(\alpha^\top x_i + \beta e_i))^{1 / \kappa_T}\\
    c_i &= \lambda_C * (-\log(1 - v_i))^{1 / \kappa_C} \quad \text{(independent case)}\\
    c_i &= \lambda_C * (-\log(1 - v_i) / \exp(\alpha^\top x_i + \beta e_i))^{1 / \kappa_C} \quad \text{(dependent case)}
\end{align*}where 
\begin{itemize}
    \item $u_i$ and $v_i$ are  drawn from two independent uniform distributions
    \item $\alpha = (1,-\exp(-1 / 10), \exp(-2 /10),0,0,0)$ and $\beta$ varies in $\{0,0.2,0.4,0.6,0.8,1.0\}$
    \item $\lambda_C, \kappa_T, \kappa_C$ have been chosen to reach a censoring level of about 15\%.
\end{itemize} 

The observed outcomes are then defined as $t_i = \min(\tau_i, c_i), 
\quad \delta_i = \mathbf{1}\{\tau_i \leq c_i\}$.

\subsection{Variable distribution in real datasets}\label{apd:distribution_real_Data}
Across all datasets, each figure consists of three panels: Kaplan–Meier survival curves of observed times (panel \textbf{A}), distribution plots of positive and continuous variables (panel \textbf{B}), and count plots of categorical and ordinal variables (panel \textbf{C}). All plots are stratified by treatment arm (control vs. treated).

\begin{figure}[!h]
    \centering
    \includegraphics[width=1.\columnwidth]{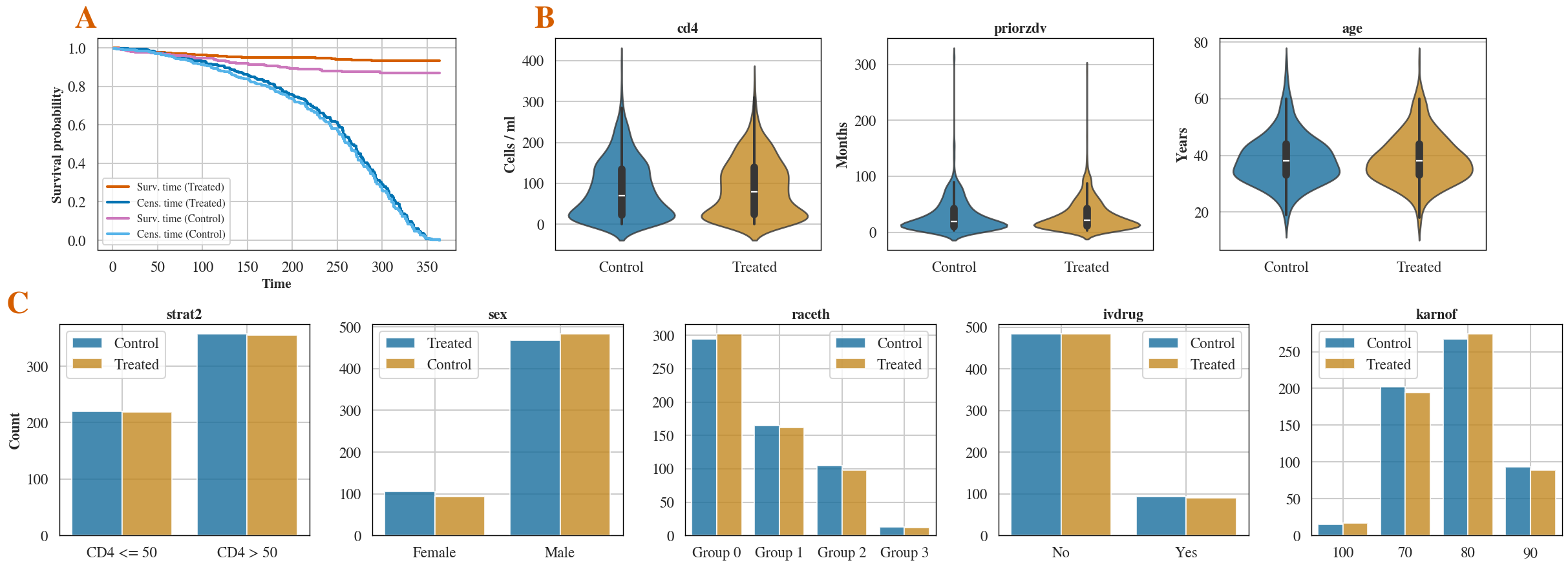}
    \caption{Variable distributions in the ACTG320 dataset.}
    \label{fig:actg320_all}
\end{figure}

\begin{figure}[!h]
    \centering
    \includegraphics[width=1.\columnwidth]{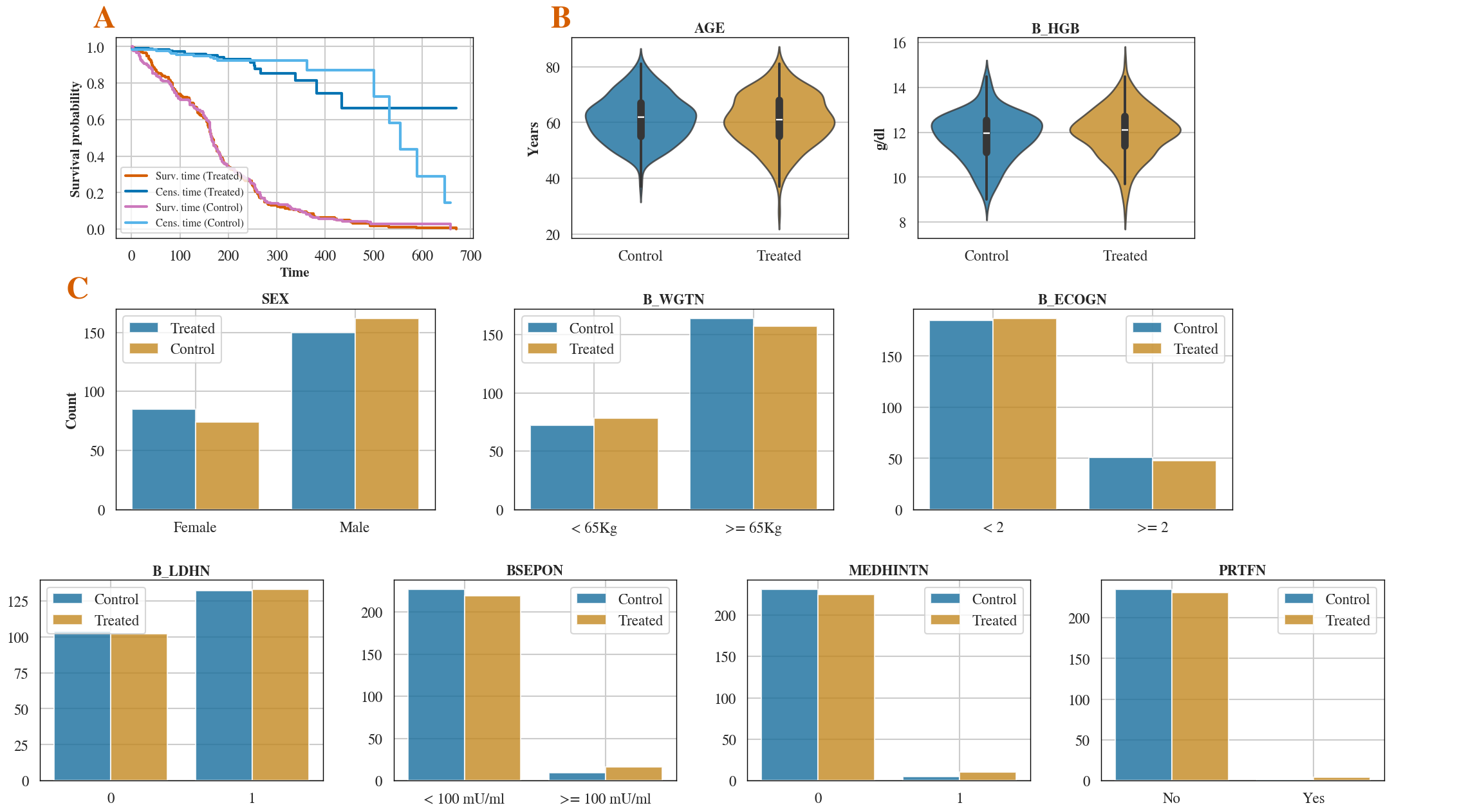}
    \caption{Variable distributions in the NCT00119613 dataset.}
    \label{fig:sas1_all}
\end{figure}

\begin{figure}[!h]
    \centering
    \includegraphics[width=1.\columnwidth]{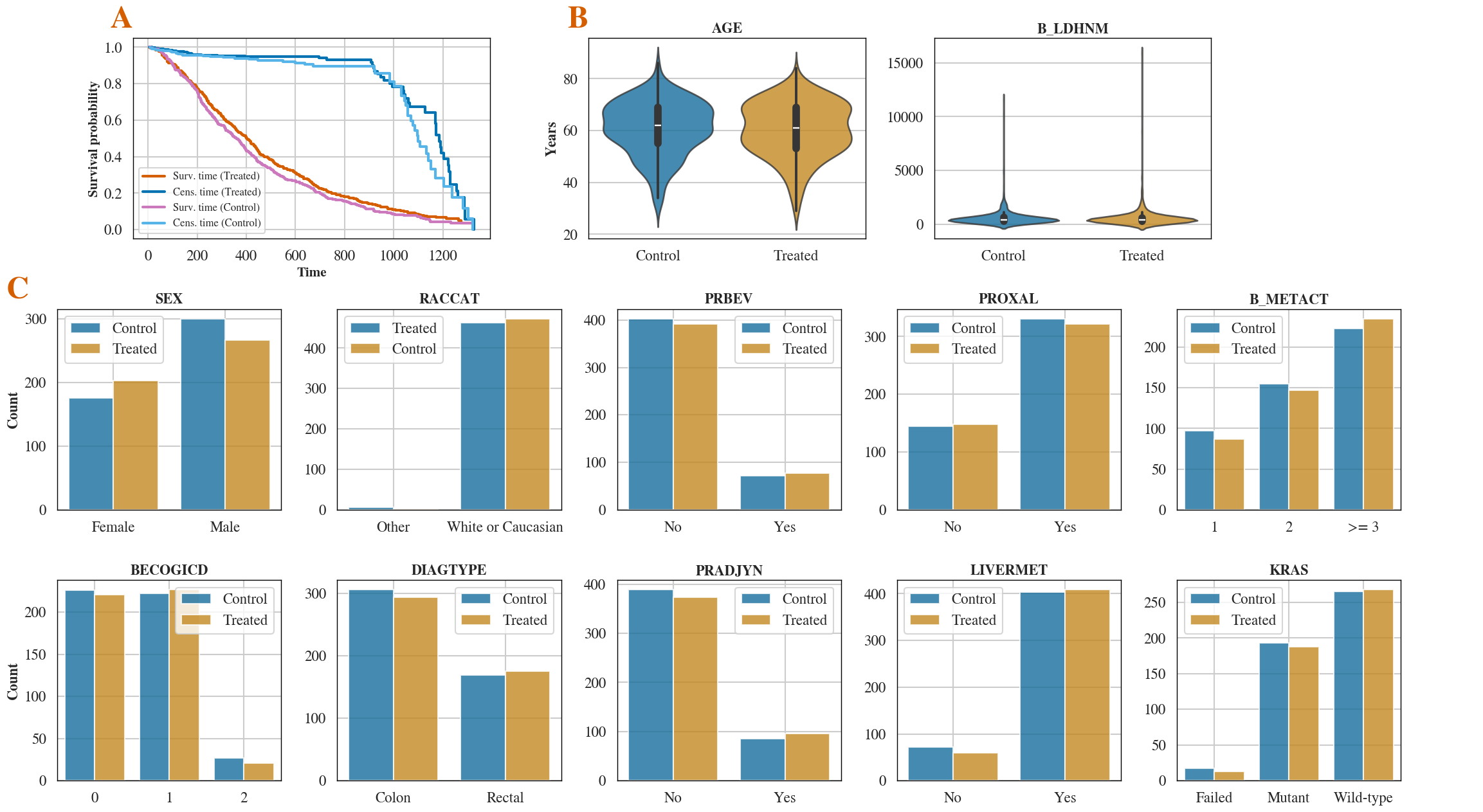}
    \caption{Variable distributions in the NCT00113763 dataset.}
    \label{fig:sas2_all}
\end{figure}

\begin{figure}[!h]
    \centering
    \includegraphics[width=1.\columnwidth]{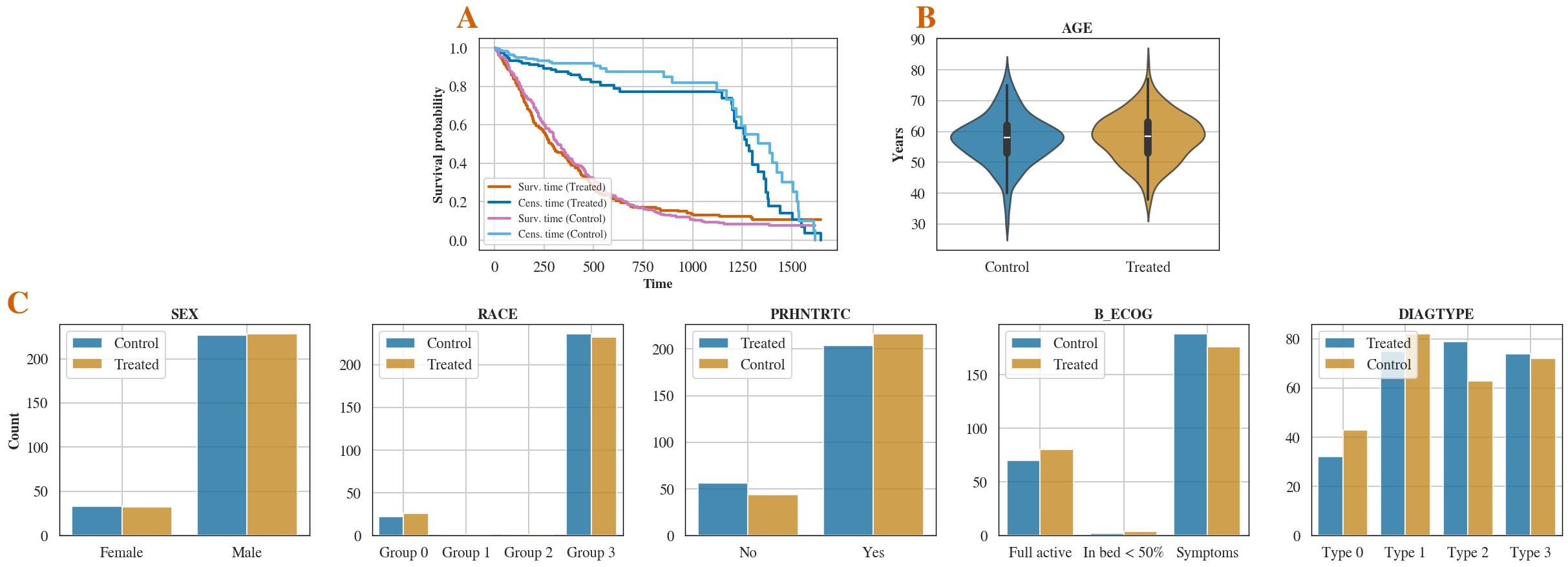}
    \caption{Variable distributions in the NCT00339183 dataset.}
    \label{fig:sas3_all}
\end{figure}

\clearpage 

\section{Supplementary materials for Section~\ref{sec:results}}

\subsection{Additional metrics} \label{apd:additional_metrics_perf}

We also evaluate the performance of our HI-VAE models against competing methods using the additional metrics described in Appendix~\ref{apd:additional_metrics},  on both simulated and real datasets.

\begin{figure}[h!]
    \centering
    \includegraphics[width=.82\linewidth]{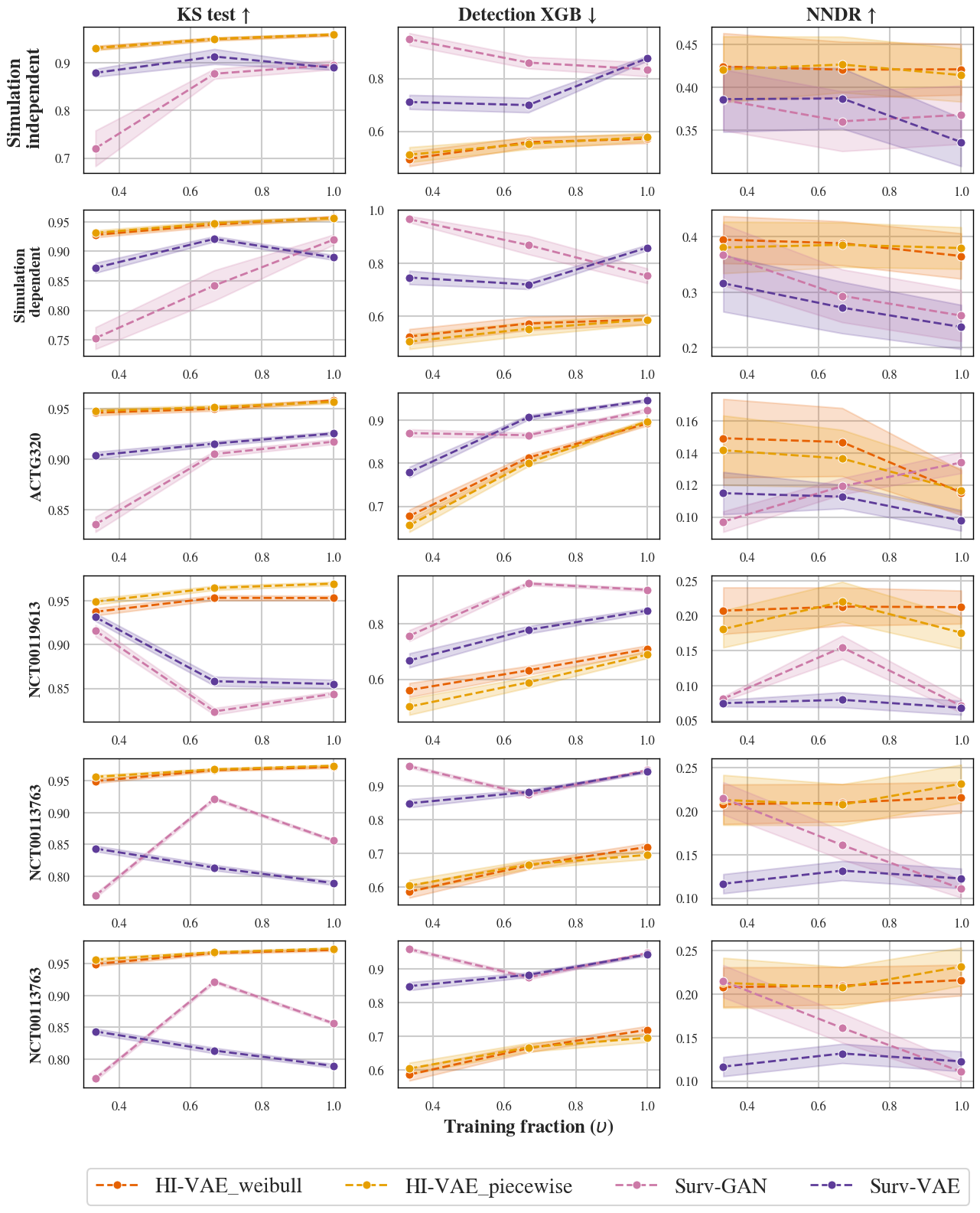}
    \caption{\footnotesize Generative performance comparison on simulated and real datasets, using KS test, detection XGB, NNDR. Arrows indicate the direction corresponding to better performance for each metric.}
    \label{fig:placeholder}
\end{figure}

\clearpage

\subsection{Post-generation selection}
\label{apd:metrics_best}

Here we present additional results on post-generation selection, as discussed in Section~\ref{subsec:post_generation_selection} of the main paper.

\begin{table}[!h]
\small
\begin{center}
\begin{tabular}{ccccc}
\hline
\multicolumn{2}{c}{\textbf{Algorithm}}&1/3& 2/3 & 1  \\ [0.7ex]
\hline
\rule{0pt}{3ex}
\multirow{4}{0.3cm}{\rotatebox{90}{\tiny{\shortstack{ (Independent) \\Simulation}}}}
& \texttt{HI-VAE\_weibull} & 0.970 & 0.768 & 0.745 \\
& \texttt{HI-VAE\_piecewise} & 0.922 & 0.842 & 0.790 \\
& \texttt{Surv-GAN}& 0.300 & 0.318 & 0.390 \\
& \texttt{Surv-VAE}& 0.540 & 0.684 & 0.280 \\
\hline
\multirow{4}{0.3cm}{\rotatebox{90}{\tiny{\shortstack{ (Dependent) \\ Simulation}}}}
& \texttt{HI-VAE\_weibull} & 0.599 & 0.628 & 0.707 \\
& \texttt{HI-VAE\_piecewise} & 0.947 & 0.491 & 0.154 \\
& \texttt{Surv-GAN} & 0.440 & 0.395 & 0.285 \\
& \texttt{Surv-VAE} & 0.252 & 0.603 & 0.454 \\
\hline
\rule{0pt}{3ex}
\multirow{4}{0.3cm}{\rotatebox{90}{\tiny{ACTG 320}}}
& \texttt{HI-VAE\_weibull} & 1.000 & 0.870 & 1.000 \\
& \texttt{HI-VAE\_piecewise} & 1.000 & 1.000 & 1.000 \\
& \texttt{Surv-GAN}& 0.970 & 1.000 & 1.000 \\
& \texttt{Surv-VAE}& 0.970 & 0.990 & 0.995 \\
\hline
\rule{0pt}{3ex}
\multirow{4}{0.3cm}{\rotatebox{90}{\tiny{NCT00119613}}}
& \texttt{HI-VAE\_weibull} & 1.000 & 0.970 & 0.990 \\
& \texttt{HI-VAE\_piecewise} & 1.000 & 1.000 & 0.990 \\
& \texttt{Surv-GAN}& 0.990 & 0.810 & 0.000 \\
& \texttt{Surv-VAE}& 0.925 & 0.960 & 0.990 \\
\hline
\rule{0pt}{3ex}
\multirow{4}{0.3cm}{\rotatebox{90}{\tiny{NCT00113763}}}
& \texttt{HI-VAE\_weibull} & 0.990 & 0.965 & 0.880 \\
& \texttt{HI-VAE\_piecewise} & 0.870 & 0.970 & 0.940 \\
& \texttt{Surv-GAN}& 1.000 & 0.000 & 0.000 \\
& \texttt{Surv-VAE}& 0.975 & 0.970 & 0.975 \\
\hline
\rule{0pt}{3ex}
\multirow{4}{0.3cm}{\rotatebox{90}{\tiny{NCT00339183}}}
& \texttt{HI-VAE\_weibull} & 1.000 & 1.000 & 0.995 \\
& \texttt{HI-VAE\_piecewise} & 1.000 & 0.995 & 1.000 \\
& \texttt{Surv-GAN}& 1.000 & 1.000 & 0.360 \\
& \texttt{Surv-VAE}& 0.915 & 0.820 & 0.990 \\
\hline
\end{tabular}
\end{center}
\caption{\footnotesize Proportion of accepted $H_0$  ($\alpha=0.05$) in log-rank tests comparing original and generated controls.} \label{tab:H0_ratio}
\end{table}

\begin{figure}[h!]
    \centering
    \includegraphics[width=1.0\columnwidth]{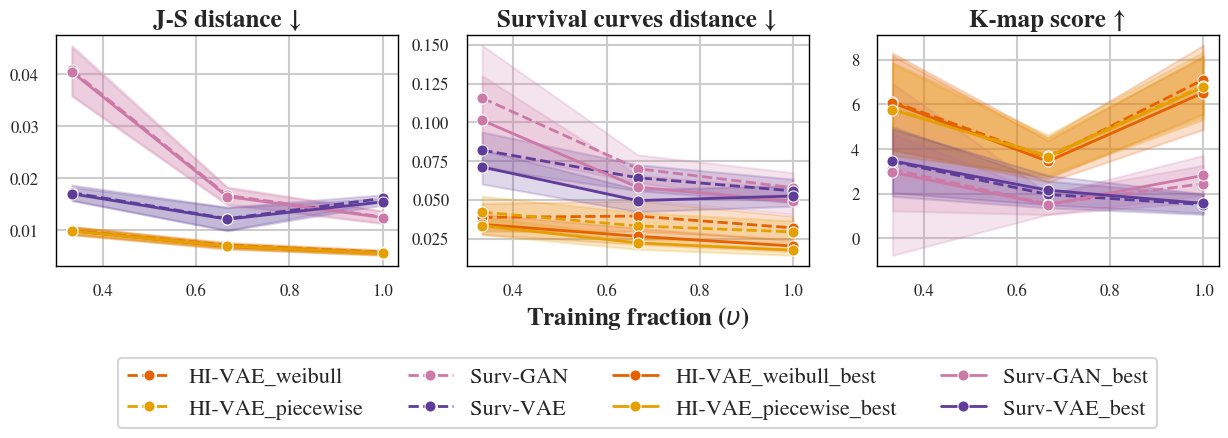}
    \caption{\footnotesize Comparison of generative performance between the best generated dataset and the full set of generated datasets across Monte Carlo experiments, under the independent simulation setting with training restricted to control-arm samples.}
    \label{fig:compare_simu_perf_control}
\end{figure}

\begin{figure}[h!]
    \centering
    \includegraphics[width=1.0\columnwidth]{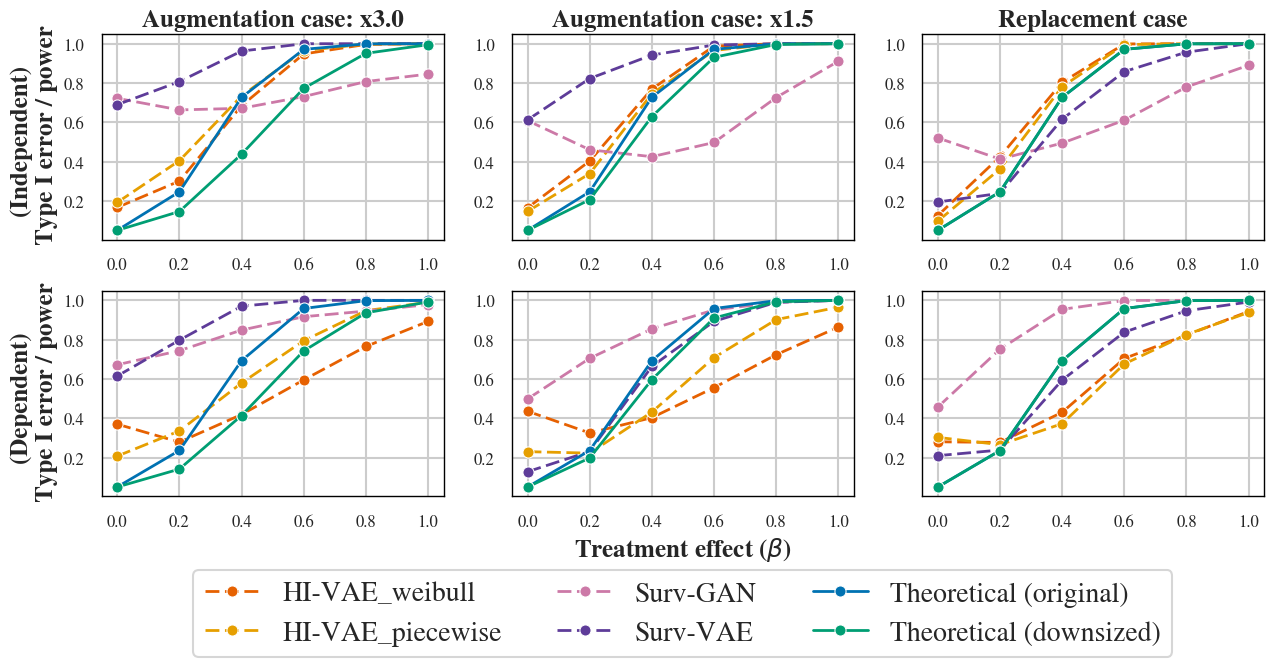}
    \caption{\footnotesize Type I error and power estimation after post-generation selection, based on subsets of the top 20\% best generated control arm, for \textbf{independent} case (\textbf{top}) and \textbf{dependent} case (\textbf{bottom}). Dashed lines: empirical power. Green: theoretical power with reduced control size. Blue: theoretical power with generated control size.}
    \label{fig:simu_pvalue_control_best_40_indep_dep}
\end{figure}

\begin{figure}[h!]
    \centering
    \includegraphics[width=1.0\columnwidth]{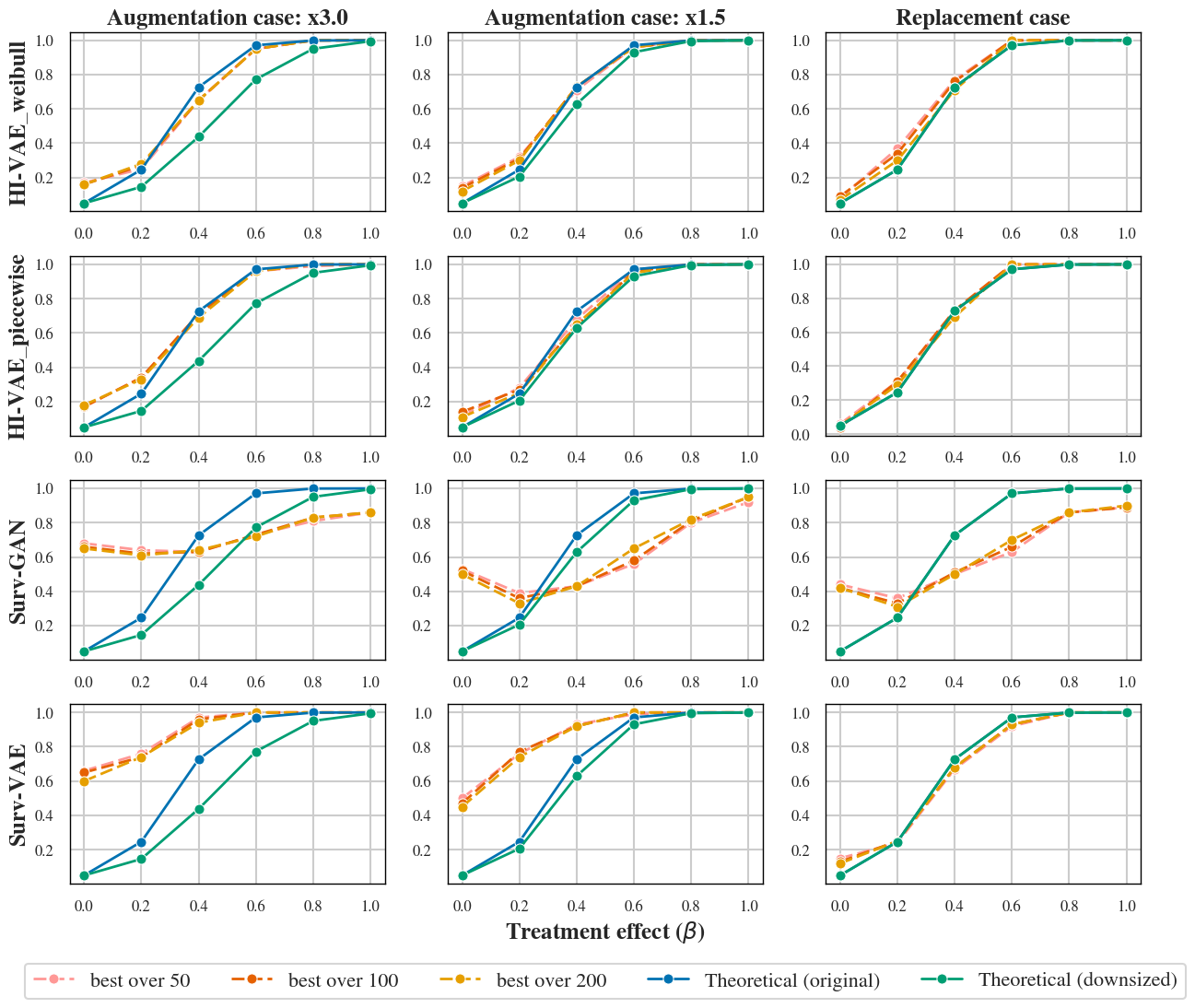}
        \caption{\footnotesize Type I error and power estimation after post-generation selection, based on subsets of varying sizes from the well-performing generated datasets, for an independent simulation setting. Dashed lines: empirical power. Green: theoretical power with reduced control size. Blue: theoretical power with generated control size.}    \label{fig:simu_perf_control_best_vs_all}
\end{figure}






\clearpage
\subsection{Risk-model discrimination/calibration}
\label{subsec:discrimination/calibration}
We report here the discrimination (C-index) and calibration (integrated Brier score) of Cox models trained separately on either real control data (using a fraction $\upsilon = 2/3$ of the available controls in the augmentation setting) or on the corresponding synthetic data. The predictive performance was then evaluated on an independent real test set consisting of the remaining control samples. Our methods yield synthetic-trained models whose performance closely aligns with that of real controls in both discrimination and calibration, whereas Surv-GAN and Surv-VAE exhibit good discrimination but poorer calibration.
\begin{figure}[h!]
    \centering
    \includegraphics[width=.7\linewidth]{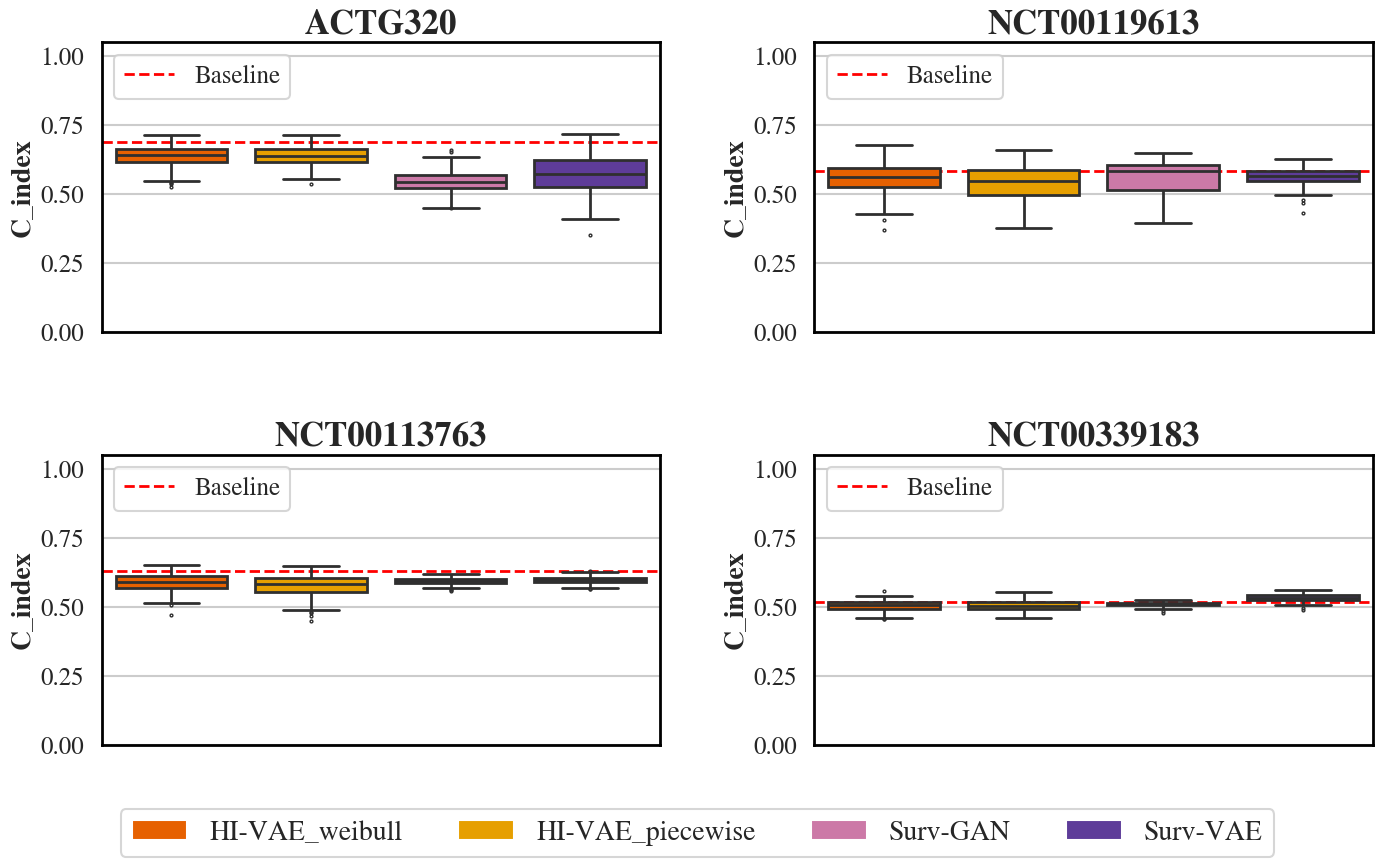}
    \caption{\footnotesize Discrimination performance comparison on real datasets, using C-index metric (\textit{higher} is better). The dashed line represents the performance of the model trained on real control data.}
    \label{fig:real-Cindex}
\end{figure}

\begin{figure}[h!]
    \centering
    \includegraphics[width=.7\linewidth]{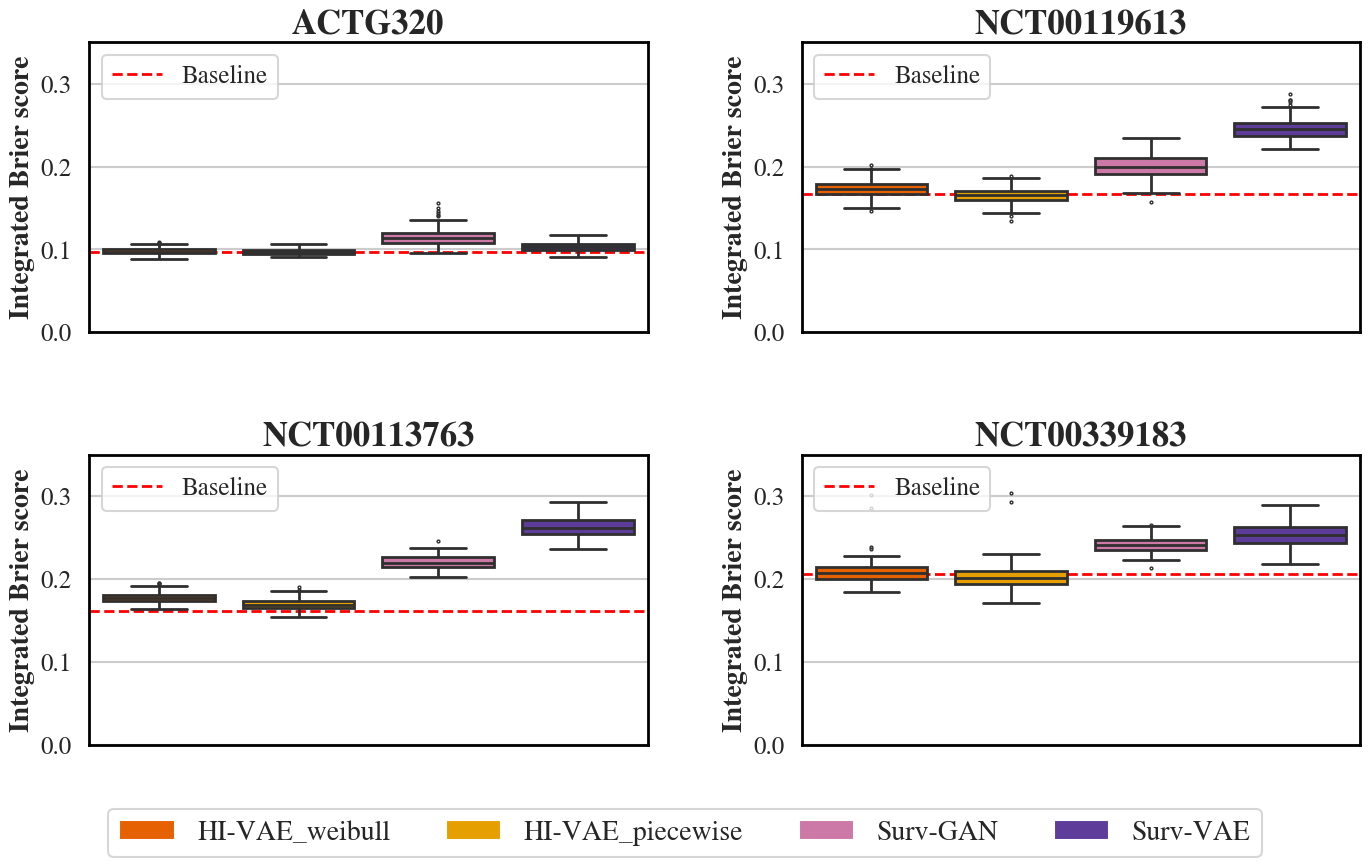}
    \caption{\footnotesize Calibration performance comparison on real datasets, using integrated Brier score metric (\textit{lower} is better). The dashed line represents the performance of the model trained on real control data.}
    \label{fig:real-BS}
\end{figure}

\clearpage

\subsection{Impact of hyperparameter optimization} \label{subsec:influence_hyperopt}

We report here the results of the additional experiments assessing the impact of both the hyperparameter search strategy and the random seed within the Optuna framework (Section~\ref{sec:extended_XP} of the main paper).

\paragraph{Influence of the chosen method for the hyperparameters search} 

We compare the following hyperparameter search methods:
\begin{enumerate}
    \item 
    Train on the full dataset of shape $N_{\text{C}}$, then generate $N_{\text{gen}}$ synthetic datasets of size $N_{\text{sim}} = N_{\text{C}}$. 
    
    \item 
    Split the dataset into training and validation sets. Train on the training set, then generate $N_{\text{gen}}$ synthetic datasets from the validation set of size $N_{\text{sim}} = N_{\text{val}}$. 

    \item 
    Split the dataset into training and validation sets. Train on the training set, then generate $N_{\text{gen}}$ synthetic datasets from the full dataset (train + validation) of size $N_{\text{sim}} = N_{\text{C}}$. 
\end{enumerate}
In all cases, the score is the survival curve distance between the generated and original control arms. For this comparison, we focus on the replacement case ($\upsilon = 1$) in the independent setting trained on controls only. 



\begin{figure}[h!]
    \centering
    \includegraphics[width=0.89\columnwidth]{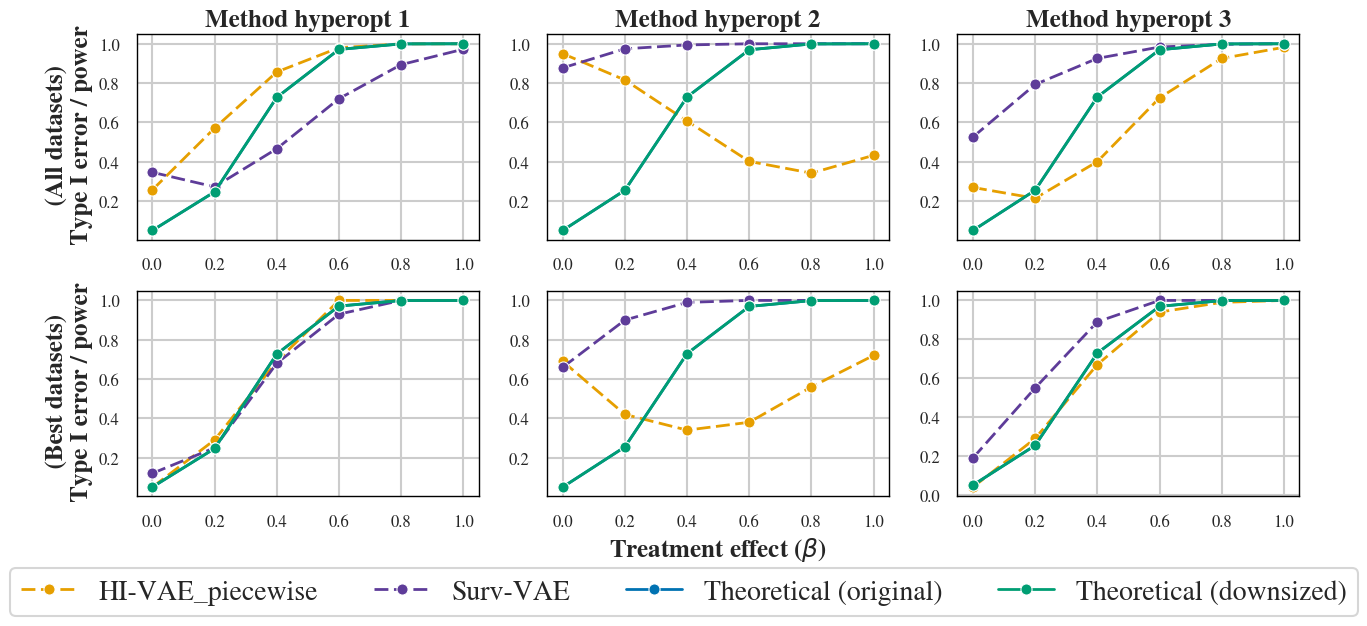}
    \caption{\footnotesize Type I error and power estimation (independent simulation setting, trained on controls only, replacement case, with \textbf{different hyperparameter search methods} and post-generation selection). Top: all generated datasets; bottom: best generated dataset. Dashed lines: empirical power. Green: theoretical power with reduced control size. Blue: theoretical power with generated control size.}
    \label{fig:simu_pvalue_comp_method_hyperopt}
\end{figure}

\begin{figure}[h!]
    \centering
    \includegraphics[width=0.85\columnwidth]{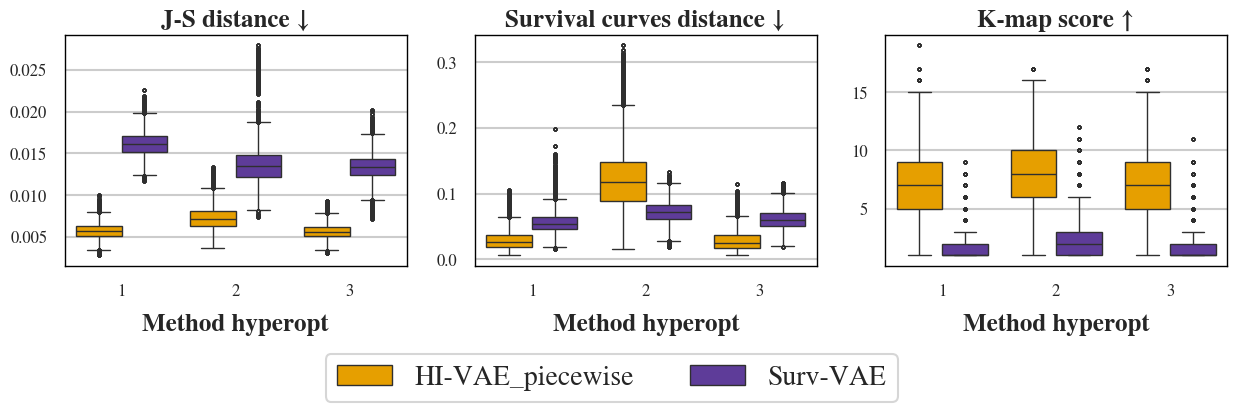}
    \caption{\footnotesize Comparison of generative performance metrics (J–S distance, survival distance, and $K$-map) across hyperparameter search methods, under the independent simulation setting (trained on controls only, replacement case).}
    \label{fig:simu_perf_comp_method_hyperopt}
\end{figure}

\paragraph{Influence of the seed in the Optuna hyperparameters search framework} We investigate the influence of the random seed in the Optuna hyperparameter search framework~\citep{akiba2019optuna}, focusing on the replacement case $\upsilon = 1$ in the independent setting trained on the control arm only, and using method~1 for the hyperparameter search.



\begin{figure}[h!]
    \centering
    \includegraphics[width=0.89\columnwidth]{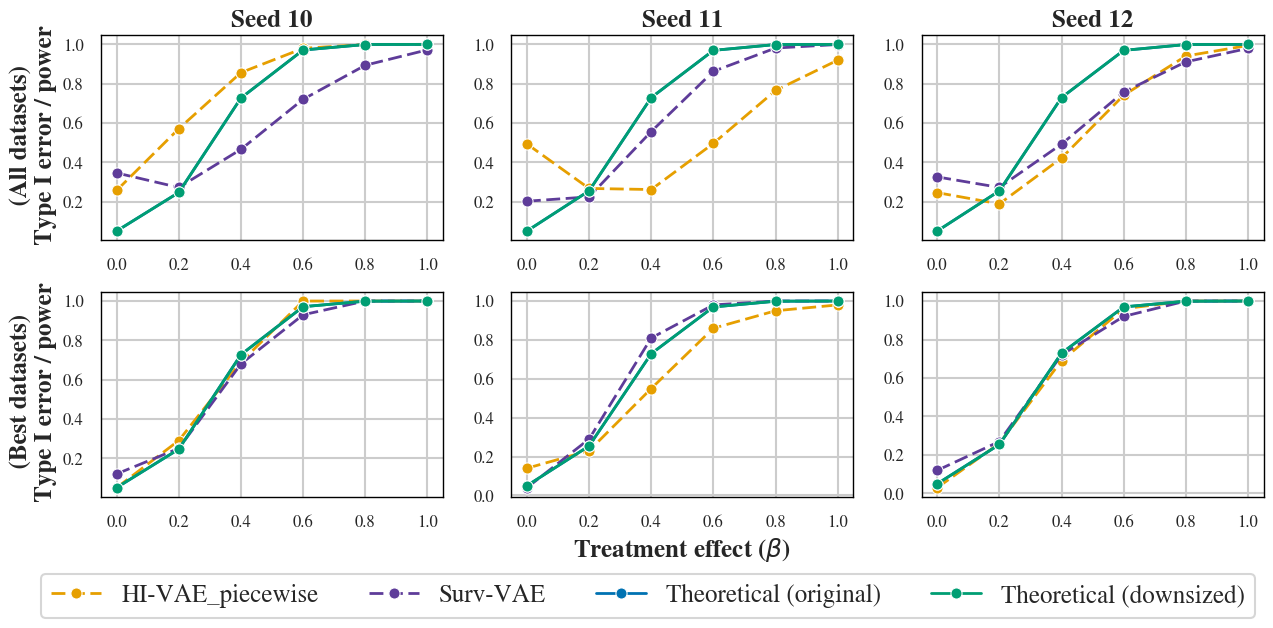}
    \caption{\footnotesize Type I error and power estimation (independent simulation setting, trained on controls only, replacement case, with \textbf{different random seed values in the hyperparameter search framework} and post-generation selection). Top: all generated datasets; bottom: best generated dataset. Dashed lines: empirical power. Green: theoretical power with reduced control size. Blue: theoretical power with generated control size.}
    \label{fig:simu_pvalue_comp_seed_hyperopt}
\end{figure}

\begin{figure}[h!]
    \centering
    \includegraphics[width=0.85\columnwidth]{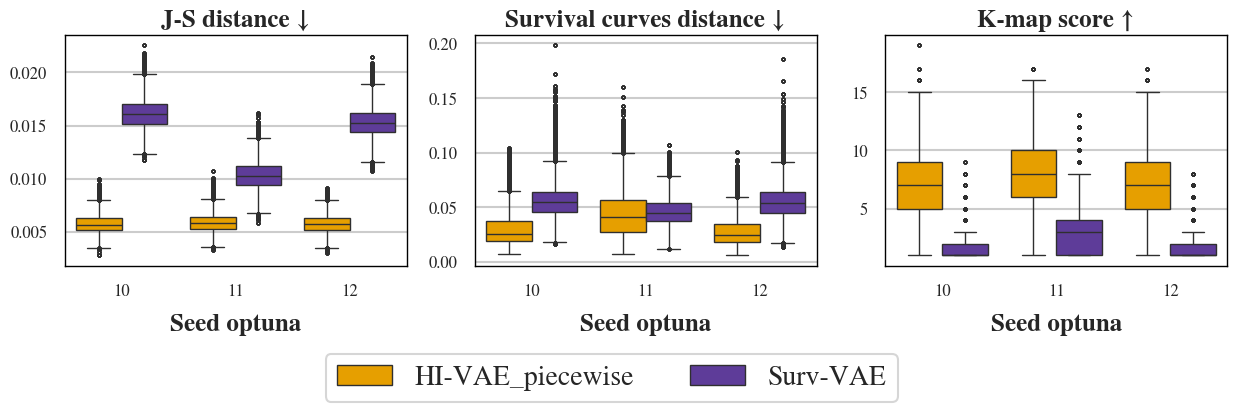}
    \caption{\footnotesize Comparison of generative performance metrics (J–S distance, survival distance, and $K$-map) across random seed values in hyperparameter search framework, in the independent simulation setting (trained on controls only, replacement case).}
    \label{fig:simu_perf_comp_seed_hyperopt}
\end{figure}

\clearpage

\subsection{Influence of the training setup: control vs. control + treated} \label{subsec:control_vs_full}

Here we provide additional results comparing models trained only on controls versus models trained on both control and treated data, as described in Section~\ref{sec:extended_XP} of the main paper.

\begin{table}[!h]
\scriptsize
\begin{center}
\caption{Comparison of generative performance metrics (J-S distance, survival distance, $K$-map) across training strategies (controls only vs. controls + treated) in the independent simulation setting.}
\begin{tabular}{cc|cc|cc|cc}
\hline
$\mathbf{\upsilon}$& \textbf{Algorithm}&\multicolumn{2}{c|}{J-S dist $\downarrow$}& \multicolumn{2}{c|}{Surv dist $\downarrow$ }& \multicolumn{2}{c}{$K$-map score $\uparrow$}  \\ [0.7ex]
& & Control & Control + Treated  & Control & Control + Treated & Control & Control + Treated  \\
\hline
\multirow{4}{0.3cm}{\rotatebox{90}{1/3}}
& \texttt{HI-VAE\_weibull} & \contour{black}{0.010±0.001} & 0.011±0.002 & \contour{black}{0.039±0.018} & 0.072±0.032 & 6.127±4.345 & \contour{black}{6.254±4.479} \\
& \texttt{HI-VAE\_piecewise} & \contour{black}{0.010±0.001} & 0.011±0.002 & \contour{black}{0.042±0.021} & 0.067±0.032 & 5.797±4.14 & \contour{black}{6.208±4.466} \\
& \texttt{Surv-GAN} & 0.041±0.01 & \contour{black}{0.021±0.006} & 0.116±0.068 & \contour{black}{0.071±0.025} & 3.091±7.692 & \contour{black}{3.541±3.338} \\
& \texttt{Surv-VAE} & 0.017±0.003 & \contour{black}{0.016±0.002} & 0.082±0.023 & \contour{black}{0.065±0.025} & 3.440±3.116 & \contour{black}{4.864±3.809} \\
\hline
\multirow{4}{0.3cm}{\rotatebox{90}{2/3}}
& \texttt{HI-VAE\_weibull} & \contour{black}{0.007±0.001} & 0.008±0.001 & \contour{black}{0.040±0.02} & 0.040±0.019 & 3.606±1.845 & \contour{black}{4.038±1.982} \\
& \texttt{HI-VAE\_piecewise} & \contour{black}{0.007±0.001} & 0.008±0.001 & \contour{black}{0.033±0.017} & 0.042±0.021 & 3.702±1.859 & \contour{black}{3.960±1.97} \\
& \texttt{Surv-GAN} & \contour{black}{0.017±0.003} & 0.022±0.008 & \contour{black}{0.070±0.018} & 0.072±0.029 & \contour{black}{1.544±0.947} & 1.373±0.762 \\
& \texttt{Surv-VAE} & \contour{black}{0.012±0.004} & 0.015±0.001 & 0.064±0.017 & \contour{black}{0.059±0.018} & \contour{black}{1.969±1.20} & 1.503±0.819 \\
\hline
\multirow{4}{0.3cm}{\rotatebox{90}{3/3}}
& \texttt{HI-VAE\_weibull} & \contour{black}{0.006±0.001} & 0.006±0.001 & \contour{black}{0.032±0.015} & 0.033±0.015 & 7.105±3.069 & \contour{black}{7.280±3.153} \\
& \texttt{HI-VAE\_piecewise} & \contour{black}{0.006±0.001} & 0.006±0.001 & \contour{black}{0.029±0.014} & 0.035±0.017 & 6.826±3.025 & \contour{black}{7.529±3.16} \\
& \texttt{Surv-GAN} & \contour{black}{0.013±0.002} & 0.032±0.006 & \contour{black}{0.058±0.019} & 0.091±0.044 & \contour{black}{2.449±1.646} & 1.237±0.615 \\
& \texttt{Surv-VAE} & 0.016±0.001 & \contour{black}{0.014±0.001} & 0.056±0.015 & \contour{black}{0.049±0.013} & 1.517±0.888 & \contour{black}{2.103±1.383} \\
\hline
\end{tabular}
\end{center}
\end{table}

\begin{table}[!h]
\scriptsize
\begin{center}
\caption{Comparison of generative performance metrics (KS test, Detect XGB, NNDR) across training strategies (controls only vs. controls + treated) in the independent simulation setting.}
\begin{tabular}{cc|cc|cc|cc}
\hline
$\mathbf{\upsilon}$& \textbf{Algorithm}&\multicolumn{2}{c|}{KS test $\uparrow$}& \multicolumn{2}{c|}{Detection XGB $\downarrow$ }& \multicolumn{2}{c}{NNDR $\uparrow$}  \\ [0.7ex]
& & Control & Control + Treated  & Control & Control + Treated & Control & Control + Treated  \\
\hline
\multirow{4}{0.3cm}{\rotatebox{90}{1/3}}
& \texttt{HI-VAE\_weibull} & \contour{black}{0.930±0.008} & 0.923±0.01 & \contour{black}{0.496±0.056} & 0.547±0.057 & 0.424±0.078 & \contour{black}{0.430±0.073} \\
& \texttt{HI-VAE\_piecewise} & \contour{black}{0.930±0.008} & 0.924±0.01 & \contour{black}{0.512±0.057} & 0.541±0.056 & 0.421±0.077 & \contour{black}{0.426±0.075} \\
& \texttt{Surv-GAN} & 0.720±0.074 & \contour{black}{0.843±0.04} & 0.950±0.046 & \contour{black}{0.840±0.078} & 0.385±0.071 & \contour{black}{0.402±0.078} \\
& \texttt{Surv-VAE} & 0.878±0.016 & \contour{black}{0.890±0.012} & \contour{black}{0.711±0.054} & 0.753±0.045 & \contour{black}{0.386±0.076} & 0.380±0.084 \\
\hline
\multirow{4}{0.3cm}{\rotatebox{90}{2/3}}
& \texttt{HI-VAE\_weibull} & \contour{black}{0.949±0.006} & 0.946±0.007 & 0.559±0.04 & \contour{black}{0.554±0.041} & 0.421±0.065 & \contour{black}{0.432±0.068} \\
& \texttt{HI-VAE\_piecewise} & \contour{black}{0.949±0.006} & 0.946±0.007 & \contour{black}{0.553±0.041} & 0.557±0.04 & 0.427±0.065 & \contour{black}{0.429±0.068} \\
& \texttt{Surv-GAN} & \contour{black}{0.876±0.021} & 0.831±0.052 & \contour{black}{0.861±0.044} & 0.921±0.047 & 0.360±0.07 & \contour{black}{0.387±0.068} \\
& \texttt{Surv-VAE} & \contour{black}{0.912±0.031} & 0.901±0.008 & \contour{black}{0.700±0.056} & 0.771±0.032 & \contour{black}{0.387±0.07} & 0.358±0.064 \\
\hline
\multirow{4}{0.3cm}{\rotatebox{90}{3/3}}
& \texttt{HI-VAE\_weibull} & \contour{black}{0.958±0.005} & 0.955±0.006 & \contour{black}{0.572±0.033} & 0.572±0.033 & \contour{black}{0.421±0.059} & 0.419±0.063 \\
& \texttt{HI-VAE\_piecewise} & \contour{black}{0.958±0.005} & 0.955±0.006 & 0.578±0.033 & \contour{black}{0.573±0.032} & 0.414±0.062 & \contour{black}{0.422±0.062} \\
& \texttt{Surv-GAN} & \contour{black}{0.895±0.019} & 0.725±0.043 & \contour{black}{0.835±0.053} & 0.993±0.009 & 0.368±0.068 & \contour{black}{0.414±0.067} \\
& \texttt{Surv-VAE} & 0.889±0.009 & \contour{black}{0.905±0.008} & 0.877±0.02 & \contour{black}{0.846±0.023} & 0.336±0.056 & \contour{black}{0.341±0.058} \\
\hline
\end{tabular}
\end{center}
\end{table}

\begin{table}[!h]
\scriptsize
\begin{center}
\caption{Comparison of generative performance metrics (J-S distance, survival distance, $K$-map) across training strategies (controls only vs. controls + treated) for the ACTG320 dataset.}
\begin{tabular}{cc|cc|cc|cc}
\hline
$\mathbf{\upsilon}$& \textbf{Algorithm}&\multicolumn{2}{c|}{J-S dist $\downarrow$}& \multicolumn{2}{c|}{Surv dist $\downarrow$ }& \multicolumn{2}{c}{$K$-map score $\uparrow$}  \\ [0.7ex]
& & Control & Control + Treated  & Control & Control + Treated & Control & Control + Treated  \\
\hline
\multirow{4}{0.3cm}{\rotatebox{90}{1/3}}
& \texttt{HI-VAE\_weibull} & \contour{black}{0.007±0.001} & 0.007±0.001 & \contour{black}{0.019±0.008} & 0.023±0.01 & 1.435±0.799 & \contour{black}{1.570±0.905} \\
& \texttt{HI-VAE\_piecewise} & \contour{black}{0.006±0.001} & 0.007±0.001 & \contour{black}{0.020±0.01} & 0.032±0.013 & \contour{black}{1.415±0.689} & 1.375±0.661 \\
& \texttt{Surv-GAN} & 0.021±0.002 & \contour{black}{0.013±0.001} & \contour{black}{0.025±0.012} & 0.059±0.012 & \contour{black}{1.325±0.584} & 1.175±0.535 \\
& \texttt{Surv-VAE} & 0.013±0.001 & \contour{black}{0.012±0.001} & \contour{black}{0.040±0.014} & 0.050±0.014 & 1.200±0.491 & \contour{black}{1.715±0.779} \\
\hline
\multirow{4}{0.3cm}{\rotatebox{90}{2/3}}
& \texttt{HI-VAE\_weibull} & 0.006±0.001 & \contour{black}{0.006±0.001} & 0.018±0.01 & \contour{black}{0.015±0.006} & \contour{black}{4.740±1.737} & 4.620±1.726 \\
& \texttt{HI-VAE\_piecewise} & \contour{black}{0.005±0.001} & 0.006±0.001 & \contour{black}{0.011±0.006} & 0.012±0.006 & \contour{black}{5.050±1.875} & 3.385±1.438 \\
& \texttt{Surv-GAN} & 0.011±0.001 & \contour{black}{0.009±0.001} & 0.012±0.005 & \contour{black}{0.009±0.001} & 1.915±1.016 & \contour{black}{2.235±1.432} \\
& \texttt{Surv-VAE} & \contour{black}{0.009±0.001} & 0.010±0.001 & \contour{black}{0.012±0.006} & 0.015±0.008 & 1.700±0.946 & \contour{black}{3.315±1.472} \\
\hline
\multirow{4}{0.3cm}{\rotatebox{90}{3/3}}
& \texttt{HI-VAE\_weibull} & \contour{black}{0.005±0.001} & 0.006±0.001 & \contour{black}{0.014±0.007} & 0.014±0.004 & 2.560±1.37 & \contour{black}{2.870±1.642} \\
& \texttt{HI-VAE\_piecewise} & \contour{black}{0.005±0.001} & 0.005±0.001 & \contour{black}{0.010±0.005} & 0.014±0.007 & \contour{black}{2.915±1.552} & 2.475±1.507 \\
& \texttt{Surv-GAN} & \contour{black}{0.010±0.001} & 0.015±0.001 & \contour{black}{0.010±0.002} & 0.012±0.002 & \contour{black}{4.790±2.046} & 2.030±1.219 \\
& \texttt{Surv-VAE} & \contour{black}{0.008±0.001} & 0.009±0.001 & \contour{black}{0.013±0.005} & 0.027±0.007 & 1.895±0.91 & \contour{black}{1.910±1.261} \\
\hline
\end{tabular}
\end{center}
\end{table}

\begin{table}[!h]
\scriptsize
\begin{center}
\caption{Comparison of generative performance metrics (KS test, Detect XGB, NNDR) across training strategies (controls only vs. controls + treated) for the ACTG320 dataset.}
\begin{tabular}{cc|cc|cc|cc}
\hline
$\mathbf{\upsilon}$& \textbf{Algorithm}&\multicolumn{2}{c|}{KS test $\uparrow$}& \multicolumn{2}{c|}{Detection XGB $\downarrow$ }& \multicolumn{2}{c}{NNDR $\uparrow$}  \\ [0.7ex]
& & Control & Control + Treated  & Control & Control + Treated & Control & Control + Treated  \\
\hline
\multirow{4}{0.3cm}{\rotatebox{90}{1/3}}
& \texttt{HI-VAE\_weibull} & \contour{black}{0.946±0.005} & 0.944±0.005 & \contour{black}{0.677±0.03} & 0.685±0.032 & \contour{black}{0.149±0.049} & 0.136±0.046 \\
& \texttt{HI-VAE\_piecewise} & \contour{black}{0.947±0.005} & 0.940±0.006 & \contour{black}{0.656±0.031} & 0.669±0.032 & \contour{black}{0.142±0.043} & 0.139±0.047 \\
& \texttt{Surv-GAN} & 0.836±0.015 & \contour{black}{0.902±0.008} & 0.870±0.017 & \contour{black}{0.813±0.021} & 0.097±0.013 & \contour{black}{0.110±0.021} \\
& \texttt{Surv-VAE} & 0.904±0.007 & \contour{black}{0.908±0.008} & 0.780±0.024 & \contour{black}{0.756±0.023} & \contour{black}{0.115±0.026} & 0.094±0.007 \\
\hline
\multirow{4}{0.3cm}{\rotatebox{90}{2/3}}
& \texttt{HI-VAE\_weibull} & 0.950±0.004 & \contour{black}{0.952±0.004} & \contour{black}{0.813±0.015} & 0.814±0.016 & \contour{black}{0.147±0.043} & 0.145±0.039 \\
& \texttt{HI-VAE\_piecewise} & 0.951±0.004 & \contour{black}{0.954±0.004} & \contour{black}{0.801±0.016} & 0.813±0.018 & \contour{black}{0.137±0.036} & 0.124±0.037 \\
& \texttt{Surv-GAN} & 0.905±0.006 & \contour{black}{0.916±0.005} & \contour{black}{0.865±0.012} & 0.880±0.011 & \contour{black}{0.119±0.01} & 0.098±0.02 \\
& \texttt{Surv-VAE} & \contour{black}{0.915±0.005} & 0.913±0.006 & 0.907±0.011 & \contour{black}{0.890±0.011} & \contour{black}{0.113±0.015} & 0.105±0.009 \\
\hline
\multirow{4}{0.3cm}{\rotatebox{90}{3/3}}
& \texttt{HI-VAE\_weibull} & \contour{black}{0.958±0.003} & 0.956±0.004 & \contour{black}{0.892±0.01} & 0.894±0.009 & 0.115±0.028 & \contour{black}{0.118±0.026} \\
& \texttt{HI-VAE\_piecewise} & 0.956±0.003 & \contour{black}{0.957±0.003} & \contour{black}{0.897±0.011} & 0.898±0.011 & \contour{black}{0.117±0.027} & 0.110±0.027 \\
& \texttt{Surv-GAN} & \contour{black}{0.917±0.005} & 0.896±0.004 & \contour{black}{0.923±0.008} & 0.946±0.005 & \contour{black}{0.134±0.013} & 0.095±0.017 \\
& \texttt{Surv-VAE} & \contour{black}{0.925±0.004} & 0.920±0.004 & 0.946±0.006 & \contour{black}{0.938±0.007} & \contour{black}{0.098±0.013} & 0.076±0.004 \\
\hline
\end{tabular}
\end{center}
\end{table}

\begin{table}[!h]
\scriptsize
\begin{center}
\caption{Comparison of generative performance metrics (J-S distance, survival distance, $K$-map) across training strategies (controls only vs. controls + treated)  for the NCT00119613 dataset.}
\begin{tabular}{cc|cc|cc|cc}
\hline
$\mathbf{\upsilon}$& \textbf{Algorithm}&\multicolumn{2}{c|}{J-S dist $\downarrow$}& \multicolumn{2}{c|}{Surv dist $\downarrow$ }& \multicolumn{2}{c}{$K$-map score $\uparrow$}  \\ [0.7ex]
& & Control & Control + Treated  & Control & Control + Treated & Control & Control + Treated  \\
\hline
\multirow{4}{0.3cm}{\rotatebox{90}{1/3}}
& \texttt{HI-VAE\_weibull} & 0.011±0.002 & \contour{black}{0.010±0.002} & 0.051±0.016 & \contour{black}{0.043±0.013} & \contour{black}{2.795±1.963} & 2.735±2.209 \\
& \texttt{HI-VAE\_piecewise} & \contour{black}{0.008±0.001} & 0.011±0.002 & \contour{black}{0.032±0.01} & 0.041±0.015 & 1.970±1.147 & \contour{black}{2.780±2.028} \\
& \texttt{Surv-GAN} & \contour{black}{0.013±0.003} & 0.024±0.002 & 0.076±0.014 & \contour{black}{0.073±0.008} & 2.890±2.789 & \contour{black}{3.405±3.662} \\
& \texttt{Surv-VAE} & \contour{black}{0.012±0.002} & 0.050±0.004 & \contour{black}{0.049±0.008} & 0.056±0.021 & \contour{black}{5.565±4.023} & 2.630±2.289 \\
\hline
\multirow{4}{0.3cm}{\rotatebox{90}{2/3}}
& \texttt{HI-VAE\_weibull} & \contour{black}{0.007±0.001} & 0.010±0.002 & \contour{black}{0.036±0.008} & 0.047±0.013 & 1.790±0.854 & \contour{black}{1.965±0.974} \\
& \texttt{HI-VAE\_piecewise} & \contour{black}{0.006±0.001} & 0.006±0.001 & 0.025±0.008 & \contour{black}{0.023±0.007} & \contour{black}{1.845±0.88} & 1.625±0.865 \\
& \texttt{Surv-GAN} & 0.026±0.001 & \contour{black}{0.016±0.002} & \contour{black}{0.048±0.006} & 0.070±0.008 & \contour{black}{1.285±0.553} & 1.255±0.549 \\
& \texttt{Surv-VAE} & \contour{black}{0.028±0.002} & 0.028±0.003 & \contour{black}{0.039±0.01} & 0.042±0.01 & 1.260±0.612 & \contour{black}{1.265±0.571} \\
\hline
\multirow{4}{0.3cm}{\rotatebox{90}{3/3}}
& \texttt{HI-VAE\_weibull} & 0.008±0.001 & \contour{black}{0.006±0.001} & 0.043±0.011 & \contour{black}{0.029±0.006} & \contour{black}{4.470±1.569} & 3.315±1.542 \\
& \texttt{HI-VAE\_piecewise} & \contour{black}{0.005±0.001} & 0.006±0.001 & \contour{black}{0.018±0.006} & 0.020±0.006 & \contour{black}{3.070±1.455} & 2.945±1.579 \\
& \texttt{Surv-GAN} & \contour{black}{0.030±0.001} & 0.038±0.001 & \contour{black}{0.038±0.007} & 0.093±0.005 & \contour{black}{1.455±0.794} & 1.060±0.356 \\
& \texttt{Surv-VAE} & \contour{black}{0.029±0.002} & 0.052±0.002 & \contour{black}{0.044±0.009} & 0.047±0.009 & \contour{black}{1.465±0.795} & 1.355±0.641 \\
\hline
\end{tabular}
\end{center}
\end{table}

\begin{table}[!h]
\scriptsize
\begin{center}
\caption{Comparison of generative performance metrics (KS test, Detect XGB, NNDR) across training strategies (controls only vs. controls + treated) for the NCT00119613 dataset.}
\begin{tabular}{cc|cc|cc|cc}
\hline
$\mathbf{\upsilon}$& \textbf{Algorithm}&\multicolumn{2}{c|}{KS test $\uparrow$}& \multicolumn{2}{c|}{Detection XGB $\downarrow$ }& \multicolumn{2}{c}{NNDR $\uparrow$}  \\ [0.7ex]
& & Control & Control + Treated  & Control & Control + Treated & Control & Control + Treated  \\
\hline
\multirow{4}{0.3cm}{\rotatebox{90}{1/3}}
& \texttt{HI-VAE\_weibull} & 0.937±0.008 & \contour{black}{0.938±0.009} & 0.561±0.051 & \contour{black}{0.556±0.054} & \contour{black}{0.207±0.066} & 0.193±0.066 \\
& \texttt{HI-VAE\_piecewise} & \contour{black}{0.949±0.007} & 0.937±0.009 & \contour{black}{0.502±0.059} & 0.542±0.062 & 0.181±0.053 & \contour{black}{0.189±0.067} \\
& \texttt{Surv-GAN} & \contour{black}{0.916±0.013} & 0.864±0.009 & \contour{black}{0.758±0.039} & 0.838±0.036 & 0.082±0.006 & \contour{black}{0.088±0.007} \\
& \texttt{Surv-VAE} & \contour{black}{0.931±0.009} & 0.757±0.018 & \contour{black}{0.669±0.051} & 0.884±0.03 & 0.075±0.009 & \contour{black}{0.079±0.007} \\
\hline
\multirow{4}{0.3cm}{\rotatebox{90}{2/3}}
& \texttt{HI-VAE\_weibull} & \contour{black}{0.953±0.006} & 0.944±0.007 & \contour{black}{0.633±0.031} & 0.635±0.036 & 0.213±0.055 & \contour{black}{0.229±0.045} \\
& \texttt{HI-VAE\_piecewise} & \contour{black}{0.965±0.005} & 0.962±0.005 & \contour{black}{0.589±0.038} & 0.599±0.036 & \contour{black}{0.220±0.057} & 0.190±0.057 \\
& \texttt{Surv-GAN} & 0.824±0.008 & \contour{black}{0.897±0.01} & 0.946±0.012 & \contour{black}{0.778±0.026} & \contour{black}{0.155±0.033} & 0.077±0.021 \\
& \texttt{Surv-VAE} & 0.858±0.011 & \contour{black}{0.861±0.012} & \contour{black}{0.779±0.027} & 0.836±0.023 & 0.080±0.022 & \contour{black}{0.080±0.027} \\
\hline
\multirow{4}{0.3cm}{\rotatebox{90}{3/3}}
& \texttt{HI-VAE\_weibull} & 0.953±0.005 & \contour{black}{0.962±0.005} & 0.710±0.026 & \contour{black}{0.694±0.025} & \contour{black}{0.213±0.047} & 0.177±0.043 \\
& \texttt{HI-VAE\_piecewise} & \contour{black}{0.970±0.004} & 0.964±0.005 & 0.689±0.026 & \contour{black}{0.688±0.028} & \contour{black}{0.176±0.045} & 0.172±0.043 \\
& \texttt{Surv-GAN} & \contour{black}{0.844±0.005} & 0.765±0.005 & \contour{black}{0.924±0.011} & 0.967±0.005 & 0.071±0.019 & \contour{black}{0.082±0.01} \\
& \texttt{Surv-VAE} & \contour{black}{0.855±0.01} & 0.746±0.011 & \contour{black}{0.848±0.018} & 0.937±0.011 & \contour{black}{0.069±0.02} & 0.062±0.009 \\
\hline
\end{tabular}
\end{center}
\end{table}

\begin{table}[!h]
\scriptsize
\begin{center}
\caption{Comparison of generative performance metrics (J-S distance, survival distance, $K$-map) across training strategies (controls only vs. controls + treated)  for the NCT00113763 dataset.}
\begin{tabular}{cc|cc|cc|cc}
\hline
$\mathbf{\upsilon}$& \textbf{Algorithm}&\multicolumn{2}{c|}{J-S dist $\downarrow$}& \multicolumn{2}{c|}{Surv dist $\downarrow$ }& \multicolumn{2}{c}{$K$-map score $\uparrow$}  \\ [0.7ex]
& & Control & Control + Treated  & Control & Control + Treated & Control & Control + Treated  \\
\hline
\multirow{4}{0.3cm}{\rotatebox{90}{1/3}}
& \texttt{HI-VAE\_weibull} & 0.009±0.001 & \contour{black}{0.008±0.001} & \contour{black}{0.025±0.008} & 0.026±0.008 & \contour{black}{2.835±1.251} & 2.530±1.173 \\
& \texttt{HI-VAE\_piecewise} & \contour{black}{0.007±0.001} & 0.008±0.001 & \contour{black}{0.039±0.017} & 0.043±0.017 & 3.385±1.279 & \contour{black}{4.370±1.346} \\
& \texttt{Surv-GAN} & 0.037±0.001 & \contour{black}{0.016±0.002} & 0.116±0.011 & \contour{black}{0.059±0.009} & 1.010±0.10 & \contour{black}{1.275±0.584} \\
& \texttt{Surv-VAE} & 0.033±0.002 & \contour{black}{0.011±0.001} & \contour{black}{0.049±0.01} & 0.062±0.01 & \contour{black}{1.435±0.691} & 1.150±0.422 \\
\hline
\multirow{4}{0.3cm}{\rotatebox{90}{2/3}}
& \texttt{HI-VAE\_weibull} & 0.006±0.001 & \contour{black}{0.006±0.001} & 0.029±0.01 & \contour{black}{0.025±0.008} & \contour{black}{9.275±2.045} & 8.545±2.136 \\
& \texttt{HI-VAE\_piecewise} & \contour{black}{0.005±0.001} & 0.007±0.001 & \contour{black}{0.025±0.01} & 0.028±0.011 & \contour{black}{10.245±1.973} & 10.175±1.855 \\
& \texttt{Surv-GAN} & \contour{black}{0.014±0.001} & 0.041±0.001 & \contour{black}{0.056±0.006} & 0.107±0.005 & \contour{black}{1.780±1.033} & 1.120±0.355 \\
& \texttt{Surv-VAE} & 0.038±0.002 & \contour{black}{0.030±0.002} & \contour{black}{0.042±0.007} & 0.046±0.007 & \contour{black}{1.840±0.969} & 1.630±0.999 \\
\hline
\multirow{4}{0.3cm}{\rotatebox{90}{3/3}}
& \texttt{HI-VAE\_weibull} & \contour{black}{0.005±0.001} & 0.005±0.001 & \contour{black}{0.025±0.009} & 0.029±0.01 & \contour{black}{13.935±2.55} & 13.875±2.773 \\
& \texttt{HI-VAE\_piecewise} & \contour{black}{0.004±0.001} & 0.005±0.001 & 0.026±0.009 & \contour{black}{0.022±0.008} & \contour{black}{14.425±2.847} & 14.27±2.674 \\
& \texttt{Surv-GAN} & \contour{black}{0.025±0.001} & 0.037±0.001 & 0.088±0.007 & \contour{black}{0.055±0.005} & 1.435±0.699 & \contour{black}{2.770±1.70} \\
& \texttt{Surv-VAE} & 0.041±0.001 & \contour{black}{0.035±0.001} & 0.044±0.007 & \contour{black}{0.041±0.007} & 1.435±0.615 & \contour{black}{2.105±1.188} \\
\hline
\end{tabular}
\end{center}
\end{table}

\begin{table}[!h]
\scriptsize
\begin{center}
\caption{Comparison of generative performance metrics (KS test, Detect XGB, NNDR) across training strategies (controls only vs. controls + treated) for the NCT00113763 dataset.}
\begin{tabular}{cc|cc|cc|cc}
\hline
$\mathbf{\upsilon}$& \textbf{Algorithm}&\multicolumn{2}{c|}{KS test $\uparrow$}& \multicolumn{2}{c|}{Detection XGB $\downarrow$ }& \multicolumn{2}{c}{NNDR $\uparrow$}  \\ [0.7ex]
& & Control & Control + Treated  & Control & Control + Treated & Control & Control + Treated  \\
\hline
\multirow{4}{0.3cm}{\rotatebox{90}{1/3}}
& \texttt{HI-VAE\_weibull} & 0.949±0.006 & \contour{black}{0.957±0.005} & \contour{black}{0.586±0.036} & 0.590±0.037 & \contour{black}{0.208±0.046} & 0.179±0.048 \\
& \texttt{HI-VAE\_piecewise} & \contour{black}{0.956±0.006} & 0.951±0.006 & \contour{black}{0.604±0.035} & 0.607±0.035 & 0.213±0.057 & \contour{black}{0.251±0.054} \\
& \texttt{Surv-GAN} & 0.770±0.007 & \contour{black}{0.916±0.008} & 0.959±0.009 & \contour{black}{0.716±0.032} & \contour{black}{0.215±0.037} & 0.127±0.022 \\
& \texttt{Surv-VAE} & 0.844±0.009 & \contour{black}{0.938±0.005} & 0.849±0.024 & \contour{black}{0.842±0.022} & 0.117±0.022 & \contour{black}{0.145±0.022} \\
\hline
\multirow{4}{0.3cm}{\rotatebox{90}{2/3}}
& \texttt{HI-VAE\_weibull} & 0.966±0.004 & \contour{black}{0.967±0.004} & 0.664±0.024 & \contour{black}{0.649±0.025} & \contour{black}{0.210±0.043} & 0.198±0.042 \\
& \texttt{HI-VAE\_piecewise} & \contour{black}{0.967±0.004} & 0.962±0.004 & 0.667±0.026 & \contour{black}{0.660±0.023} & 0.208±0.047 & \contour{black}{0.216±0.052} \\
& \texttt{Surv-GAN} & \contour{black}{0.922±0.005} & 0.792±0.004 & \contour{black}{0.875±0.012} & 0.971±0.005 & \contour{black}{0.161±0.034} & 0.140±0.022 \\
& \texttt{Surv-VAE} & 0.814±0.009 & \contour{black}{0.855±0.008} & 0.882±0.014 & \contour{black}{0.876±0.013} & \contour{black}{0.132±0.022} & 0.124±0.02 \\
\hline
\multirow{4}{0.3cm}{\rotatebox{90}{3/3}}
& \texttt{HI-VAE\_weibull} & \contour{black}{0.971±0.003} & 0.971±0.003 & 0.719±0.022 & \contour{black}{0.703±0.023} & \contour{black}{0.216±0.035} & 0.211±0.045 \\
& \texttt{HI-VAE\_piecewise} & \contour{black}{0.973±0.003} & 0.973±0.003 & \contour{black}{0.695±0.025} & 0.700±0.02 & \contour{black}{0.232±0.044} & 0.211±0.043 \\
& \texttt{Surv-GAN} & \contour{black}{0.856±0.004} & 0.828±0.004 & 0.947±0.006 & \contour{black}{0.944±0.006} & 0.111±0.021 & \contour{black}{0.118±0.016} \\
& \texttt{Surv-VAE} & 0.790±0.006 & \contour{black}{0.833±0.007} & 0.942±0.006 & \contour{black}{0.890±0.01} & \contour{black}{0.123±0.023} & 0.116±0.014 \\
\hline
\end{tabular}
\end{center}
\end{table}

\begin{table}[!h]
\scriptsize
\begin{center}
\caption{Comparison of generative performance metrics (J-S distance, survival distance, $K$-map) across training strategies (controls only vs. controls + treated)  for the NCT00339183 dataset.}
\begin{tabular}{cc|cc|cc|cc}
\hline
$\mathbf{\upsilon}$& \textbf{Algorithm}&\multicolumn{2}{c|}{J-S dist $\downarrow$}& \multicolumn{2}{c|}{Surv dist $\downarrow$ }& \multicolumn{2}{c}{$K$-map score $\uparrow$}  \\ [0.7ex]
& & Control & Control + Treated  & Control & Control + Treated & Control & Control + Treated  \\
\hline
\multirow{4}{0.3cm}{\rotatebox{90}{1/3}}
& \texttt{HI-VAE\_weibull} & \contour{black}{0.007±0.001} & 0.009±0.002 & 0.056±0.023 & \contour{black}{0.047±0.018} & \contour{black}{7.155±2.122} & 5.455±1.989 \\
& \texttt{HI-VAE\_piecewise} & \contour{black}{0.007±0.001} & 0.008±0.002 & 0.049±0.023 & \contour{black}{0.044±0.019} & \contour{black}{7.475±2.062} & 5.155±1.838 \\
& \texttt{Surv-GAN} & \contour{black}{0.010±0.001} & 0.015±0.002 & \contour{black}{0.065±0.021} & 0.138±0.017 & 2.515±1.91 & \contour{black}{3.025±2.939} \\
& \texttt{Surv-VAE} & \contour{black}{0.010±0.002} & 0.014±0.002 & 0.065±0.022 & \contour{black}{0.054±0.018} & 2.380±1.655 & \contour{black}{2.710±2.128} \\
\hline
\multirow{4}{0.3cm}{\rotatebox{90}{2/3}}
& \texttt{HI-VAE\_weibull} & \contour{black}{0.005±0.001} & 0.007±0.001 & \contour{black}{0.038±0.013} & 0.049±0.018 & 2.165±1.016 & \contour{black}{2.185±1.042} \\
& \texttt{HI-VAE\_piecewise} & \contour{black}{0.004±0.001} & 0.006±0.001 & \contour{black}{0.025±0.01} & 0.027±0.012 & \contour{black}{2.045±0.846} & 1.825±0.859 \\
& \texttt{Surv-GAN} & 0.022±0.001 & \contour{black}{0.013±0.001} & 0.121±0.01 & \contour{black}{0.082±0.012} & \contour{black}{1.570±0.938} & 1.075±0.282 \\
& \texttt{Surv-VAE} & \contour{black}{0.009±0.001} & 0.021±0.002 & 0.069±0.014 & \contour{black}{0.049±0.015} & \contour{black}{1.390±0.686} & 1.120±0.383 \\
\hline
\multirow{4}{0.3cm}{\rotatebox{90}{3/3}}
& \texttt{HI-VAE\_weibull} & \contour{black}{0.005±0.001} & 0.005±0.001 & \contour{black}{0.034±0.01} & 0.039±0.011 & \contour{black}{2.490±1.173} & 2.400±1.032 \\
& \texttt{HI-VAE\_piecewise} & \contour{black}{0.004±0.001} & 0.004±0.001 & \contour{black}{0.022±0.009} & 0.023±0.01 & \contour{black}{2.720±1.13} & 2.645±1.089 \\
& \texttt{Surv-GAN} & \contour{black}{0.012±0.001} & 0.025±0.001 & \contour{black}{0.062±0.01} & 0.067±0.008 & 1.155±0.415 & \contour{black}{1.310±0.817} \\
& \texttt{Surv-VAE} & 0.011±0.001 & \contour{black}{0.010±0.001} & 0.041±0.012 & \contour{black}{0.040±0.013} & \contour{black}{1.185±0.502} & 1.125±0.425 \\
\hline
\end{tabular}
\end{center}
\end{table}

\begin{table}[!h]
\scriptsize
\begin{center}
\caption{Comparison of generative performance metrics (KS test, Detect XGB, NNDR) across training strategies (controls only vs. controls + treated) for the NCT00339183 dataset.}
\begin{tabular}{cc|cc|cc|cc}
\hline
$\mathbf{\upsilon}$& \textbf{Algorithm}&\multicolumn{2}{c|}{KS test $\uparrow$}& \multicolumn{2}{c|}{Detection XGB $\downarrow$ }& \multicolumn{2}{c}{NNDR $\uparrow$}  \\ [0.7ex]
& & Control & Control + Treated  & Control & Control + Treated & Control & Control + Treated  \\
\hline
\multirow{4}{0.3cm}{\rotatebox{90}{1/3}}
& \texttt{HI-VAE\_weibull} & \contour{black}{0.957±0.007} & 0.948±0.008 & \contour{black}{0.526±0.046} & 0.541±0.053 & \contour{black}{0.171±0.057} & 0.132±0.032 \\
& \texttt{HI-VAE\_piecewise} & \contour{black}{0.955±0.007} & 0.950±0.008 & \contour{black}{0.519±0.05} & 0.534±0.052 & \contour{black}{0.189±0.055} & 0.156±0.039 \\
& \texttt{Surv-GAN} & \contour{black}{0.935±0.008} & 0.887±0.01 & \contour{black}{0.679±0.048} & 0.766±0.037 & \contour{black}{0.113±0.021} & 0.109±0.009 \\
& \texttt{Surv-VAE} & \contour{black}{0.937±0.009} & 0.918±0.009 & \contour{black}{0.620±0.049} & 0.723±0.043 & \contour{black}{0.110±0.02} & 0.108±0.023 \\
\hline
\multirow{4}{0.3cm}{\rotatebox{90}{2/3}}
& \texttt{HI-VAE\_weibull} & \contour{black}{0.967±0.005} & 0.958±0.006 & \contour{black}{0.676±0.037} & 0.676±0.033 & \contour{black}{0.124±0.047} & 0.116±0.037 \\
& \texttt{HI-VAE\_piecewise} & \contour{black}{0.969±0.005} & 0.963±0.005 & 0.658±0.034 & \contour{black}{0.654±0.039} & 0.127±0.037 & \contour{black}{0.130±0.037} \\
& \texttt{Surv-GAN} & 0.855±0.008 & \contour{black}{0.927±0.006} & 0.855±0.022 & \contour{black}{0.811±0.024} & \contour{black}{0.069±0.012} & 0.068±0.011 \\
& \texttt{Surv-VAE} & \contour{black}{0.945±0.006} & 0.895±0.008 & \contour{black}{0.809±0.028} & 0.810±0.026 & 0.065±0.014 & \contour{black}{0.081±0.023} \\
\hline
\multirow{4}{0.3cm}{\rotatebox{90}{3/3}}
& \texttt{HI-VAE\_weibull} & \contour{black}{0.969±0.004} & 0.967±0.005 & \contour{black}{0.803±0.026} & 0.805±0.023 & \contour{black}{0.125±0.037} & 0.123±0.036 \\
& \texttt{HI-VAE\_piecewise} & \contour{black}{0.973±0.004} & 0.972±0.005 & 0.805±0.022 & \contour{black}{0.797±0.022} & 0.135±0.039 & \contour{black}{0.137±0.033} \\
& \texttt{Surv-GAN} & \contour{black}{0.910±0.005} & 0.846±0.006 & 0.903±0.016 & \contour{black}{0.892±0.013} & \contour{black}{0.069±0.022} & 0.057±0.01 \\
& \texttt{Surv-VAE} & 0.936±0.007 & \contour{black}{0.940±0.006} & \contour{black}{0.852±0.018} & 0.872±0.018 & 0.060±0.018 & \contour{black}{0.063±0.019} \\
\hline
\end{tabular}
\end{center}
\end{table}

\clearpage

\subsection{Differences with prior vs. posterior samplings} \label{subsec:prior_vs_posterior}

The following tables illustrate the comparison between prior- and posterior-based sampling for generating synthetic data, introduced in Section~\ref{sec:extended_XP} of the main paper.

\begin{table}[!h]
\scriptsize
\begin{center}
\caption{Comparison of generative performance metrics (J-S distance, survival distance, $K$-map) across sampling strategies (posterior vs. prior)  in the independent simulation setting (trained on controls only).}
\begin{tabular}{cc|cc|cc|cc}
\hline
$\mathbf{\upsilon}$& \textbf{Algorithm}&\multicolumn{2}{c|}{J-S dist $\downarrow$}& \multicolumn{2}{c|}{Surv dist $\downarrow$ }& \multicolumn{2}{c}{$K$-map score $\uparrow$}  \\ [0.7ex]
& & Posterior & Prior & Posterior & Prior & Posterior & Prior  \\
\hline
\multirow{2}{0.3cm}{\rotatebox{90}{1/3}}
& \texttt{HI-VAE\_piecewise} & \contour{black}{0.008±0.002} & 0.008±0.002 & \contour{black}{0.035±0.018} & 0.036±0.021 & \contour{black}{5.442±3.407} & 4.974±3.174 \\
& \texttt{HI-VAE\_weibull} & \contour{black}{0.008±0.002} & 0.008±0.003 & \contour{black}{0.037±0.018} & 0.058±0.042 & \contour{black}{5.612±3.569} & 5.027±3.619 \\
\hline
\multirow{2}{0.3cm}{\rotatebox{90}{2/3}}
& \texttt{HI-VAE\_piecewise} & \contour{black}{0.008±0.002} & 0.008±0.002 & \contour{black}{0.035±0.018} & 0.036±0.021 & \contour{black}{5.442±3.407} & 4.974±3.174 \\
& \texttt{HI-VAE\_weibull} & \contour{black}{0.008±0.002} & 0.008±0.003 & \contour{black}{0.037±0.018} & 0.058±0.042 & \contour{black}{5.612±3.569} & 5.027±3.619 \\
\hline
\multirow{2}{0.3cm}{\rotatebox{90}{3/3}}
& \texttt{HI-VAE\_piecewise} & \contour{black}{0.008±0.002} & 0.008±0.002 & \contour{black}{0.035±0.018} & 0.036±0.021 & \contour{black}{5.442±3.407} & 4.974±3.174 \\
& \texttt{HI-VAE\_weibull} & \contour{black}{0.008±0.002} & 0.008±0.003 & \contour{black}{0.037±0.018} & 0.058±0.042 & \contour{black}{5.612±3.569} & 5.027±3.619 \\
\hline
\end{tabular}
\end{center}
\end{table}

\begin{table}[!h]
\scriptsize
\begin{center}
\caption{Comparison of generative performance metrics (KS test, Detect XGB, NNDR) across sampling strategies (posterior vs. prior)  in the independent simulation setting (trained on controls only)}
\begin{tabular}{cc|cc|cc|cc}
\hline
$\mathbf{\upsilon}$& \textbf{Algorithm}&\multicolumn{2}{c|}{KS test $\uparrow$}& \multicolumn{2}{c|}{Detection XGB $\downarrow$ }& \multicolumn{2}{c}{NNDR $\uparrow$}  \\ [0.7ex]
& & Posterior & Prior & Posterior & Prior & Posterior & Prior  \\
\hline
\multirow{2}{0.3cm}{\rotatebox{90}{1/3}}
& \texttt{HI-VAE\_piecewise} & \contour{black}{0.946±0.014} & 0.944±0.014 & \contour{black}{0.548±0.052} & 0.560±0.051 & \contour{black}{0.421±0.068} & 0.413±0.067 \\
& \texttt{HI-VAE\_weibull} & \contour{black}{0.946±0.013} & 0.941±0.017 & \contour{black}{0.542±0.055} & 0.584±0.048 & \contour{black}{0.422±0.068} & 0.410±0.065 \\
\hline
\multirow{2}{0.3cm}{\rotatebox{90}{2/3}}
& \texttt{HI-VAE\_piecewise} & \contour{black}{0.946±0.014} & 0.944±0.014 & \contour{black}{0.548±0.052} & 0.560±0.051 & \contour{black}{0.421±0.068} & 0.413±0.067 \\
& \texttt{HI-VAE\_weibull} & \contour{black}{0.946±0.013} & 0.941±0.017 & \contour{black}{0.542±0.055} & 0.584±0.048 & \contour{black}{0.422±0.068} & 0.410±0.065 \\
\hline
\multirow{2}{0.3cm}{\rotatebox{90}{3/3}}
& \texttt{HI-VAE\_piecewise} & \contour{black}{0.946±0.014} & 0.944±0.014 & \contour{black}{0.548±0.052} & 0.560±0.051 & \contour{black}{0.421±0.068} & 0.413±0.067 \\
& \texttt{HI-VAE\_weibull} & \contour{black}{0.946±0.013} & 0.941±0.017 & \contour{black}{0.542±0.055} & 0.584±0.048 & \contour{black}{0.422±0.068} & 0.410±0.065 \\
\hline
\end{tabular}
\end{center}
\end{table}

\begin{table}[!h]
\scriptsize
\begin{center}
\caption{Comparison of generative performance metrics (J-S distance, survival distance, $K$-map) across sampling strategies (posterior vs. prior) for the ACTG320 dataset (trained on controls only).}
\begin{tabular}{cc|cc|cc|cc}
\hline
$\mathbf{\upsilon}$& \textbf{Algorithm}&\multicolumn{2}{c|}{J-S dist $\downarrow$}& \multicolumn{2}{c|}{Surv dist $\downarrow$ }& \multicolumn{2}{c}{$K$-map score $\uparrow$}  \\ [0.7ex]
& & Posterior & Prior & Posterior & Prior & Posterior & Prior  \\
\hline
\multirow{2}{0.3cm}{\rotatebox{90}{1/3}}
& \texttt{HI-VAE\_piecewise} & 0.006±0.001 & \contour{black}{0.006±0.001} & \contour{black}{0.020±0.01} & 0.020±0.008 & 1.415±0.689 & \contour{black}{1.445±0.755} \\
& \texttt{HI-VAE\_weibull} & \contour{black}{0.007±0.001} & 0.007±0.001 & 0.019±0.008 & \contour{black}{0.018±0.008} & \contour{black}{1.435±0.799} & 1.365±0.731 \\
\hline
\multirow{2}{0.3cm}{\rotatebox{90}{2/3}}
& \texttt{HI-VAE\_piecewise} & \contour{black}{0.005±0.001} & 0.007±0.001 & \contour{black}{0.011±0.006} & 0.014±0.009 & 5.050±1.875 & \contour{black}{5.115±1.666} \\
& \texttt{HI-VAE\_weibull} & \contour{black}{0.006±0.001} & 0.007±0.001 & 0.018±0.01 & \contour{black}{0.014±0.006} & \contour{black}{4.740±1.737} & 4.045±1.714 \\
\hline
\multirow{2}{0.3cm}{\rotatebox{90}{3/3}}
& \texttt{HI-VAE\_piecewise} & \contour{black}{0.005±0.001} & 0.005±0.001 & 0.010±0.005 & \contour{black}{0.009±0.006} & \contour{black}{2.915±1.552} & 2.675±1.49 \\
& \texttt{HI-VAE\_weibull} & \contour{black}{0.005±0.001} & 0.006±0.00 & 0.014±0.007 & \contour{black}{0.012±0.005} & \contour{black}{2.560±1.37} & 2.220±1.364 \\
\hline
\end{tabular}
\end{center}
\end{table}

\begin{table}[!h]
\scriptsize
\begin{center}
\caption{Comparison of generative performance metrics (KS test, Detect XGB, NNDR) across sampling strategies (posterior vs. prior) for the ACTG320 dataset (trained on controls only).}
\begin{tabular}{cc|cc|cc|cc}
\hline
$\mathbf{\upsilon}$& \textbf{Algorithm}&\multicolumn{2}{c|}{KS test $\uparrow$}& \multicolumn{2}{c|}{Detection XGB $\downarrow$ }& \multicolumn{2}{c}{NNDR $\uparrow$}  \\ [0.7ex]
& & Posterior & Prior & Posterior & Prior & Posterior & Prior  \\
\hline
\multirow{2}{0.3cm}{\rotatebox{90}{1/3}}
& \texttt{HI-VAE\_piecewise} & \contour{black}{0.947±0.005} & 0.947±0.005 & \contour{black}{0.656±0.031} & 0.677±0.032 & \contour{black}{0.142±0.043} & 0.128±0.036 \\
& \texttt{HI-VAE\_weibull} & \contour{black}{0.946±0.005} & 0.940±0.006 & \contour{black}{0.677±0.03} & 0.707±0.03 & \contour{black}{0.149±0.049} & 0.144±0.038 \\
\hline
\multirow{2}{0.3cm}{\rotatebox{90}{2/3}}
& \texttt{HI-VAE\_piecewise} & \contour{black}{0.951±0.004} & 0.944±0.004 & \contour{black}{0.801±0.016} & 0.812±0.016 & 0.137±0.036 & \contour{black}{0.144±0.042} \\
& \texttt{HI-VAE\_weibull} & \contour{black}{0.950±0.004} & 0.948±0.004 & \contour{black}{0.813±0.015} & 0.839±0.015 & \contour{black}{0.147±0.043} & 0.135±0.043 \\
\hline
\multirow{2}{0.3cm}{\rotatebox{90}{3/3}}
& \texttt{HI-VAE\_piecewise} & \contour{black}{0.956±0.003} & 0.954±0.004 & 0.897±0.011 & \contour{black}{0.893±0.011} & \contour{black}{0.117±0.027} & 0.109±0.023 \\
& \texttt{HI-VAE\_weibull} & \contour{black}{0.958±0.003} & 0.956±0.003 & \contour{black}{0.892±0.01} & 0.905±0.009 & \contour{black}{0.115±0.028} & 0.101±0.025 \\
\hline
\end{tabular}
\end{center}
\end{table}

\begin{table}[!h]
\scriptsize
\begin{center}
\caption{Comparison of generative performance metrics (J-S distance, survival distance, $K$-map) across sampling strategies (posterior vs. prior) for the NCT00119613 dataset (trained on controls only).}
\begin{tabular}{cc|cc|cc|cc}
\hline
$\mathbf{\upsilon}$& \textbf{Algorithm}&\multicolumn{2}{c|}{J-S dist $\downarrow$}& \multicolumn{2}{c|}{Surv dist $\downarrow$ }& \multicolumn{2}{c}{$K$-map score $\uparrow$}  \\ [0.7ex]
& & Posterior & Prior & Posterior & Prior & Posterior & Prior  \\
\hline
\multirow{2}{0.3cm}{\rotatebox{90}{1/3}}
& \texttt{HI-VAE\_piecewise} & \contour{black}{0.008±0.001} & 0.011±0.002 & \contour{black}{0.032±0.01} & 0.056±0.024 & 1.970±1.147 & \contour{black}{3.185±2.686} \\
& \texttt{HI-VAE\_weibull} & 0.011±0.002 & \contour{black}{0.009±0.001} & 0.051±0.016 & \contour{black}{0.040±0.011} & \contour{black}{2.795±1.963} & 2.790±2.133 \\
\hline
\multirow{2}{0.3cm}{\rotatebox{90}{2/3}}
& \texttt{HI-VAE\_piecewise} & \contour{black}{0.006±0.001} & 0.006±0.001 & \contour{black}{0.025±0.008} & 0.030±0.011 & 1.845±0.88 & \contour{black}{1.975±0.932} \\
& \texttt{HI-VAE\_weibull} & \contour{black}{0.007±0.001} & 0.007±0.001 & \contour{black}{0.036±0.008} & 0.038±0.008 & 1.790±0.854 & \contour{black}{2.090±0.936} \\
\hline
\multirow{2}{0.3cm}{\rotatebox{90}{3/3}}
& \texttt{HI-VAE\_piecewise} & \contour{black}{0.005±0.001} & 0.005±0.001 & \contour{black}{0.018±0.006} & 0.022±0.008 & 3.070±1.455 & \contour{black}{3.490±1.701} \\
& \texttt{HI-VAE\_weibull} & 0.008±0.001 & \contour{black}{0.007±0.001} & \contour{black}{0.043±0.011} & 0.044±0.01 & 4.470±1.569 & \contour{black}{4.865±1.526} \\
\hline
\end{tabular}
\end{center}
\end{table}

\begin{table}[!h]
\scriptsize
\begin{center}
\caption{Comparison of generative performance metrics (KS test, Detect XGB, NNDR) across sampling strategies (posterior vs. prior) for the NCT00119613 dataset (trained on controls only).}
\begin{tabular}{cc|cc|cc|cc}
\hline
$\mathbf{\upsilon}$& \textbf{Algorithm}&\multicolumn{2}{c|}{KS test $\uparrow$}& \multicolumn{2}{c|}{Detection XGB $\downarrow$ }& \multicolumn{2}{c}{NNDR $\uparrow$}  \\ [0.7ex]
& & Posterior & Prior & Posterior & Prior & Posterior & Prior  \\
\hline
\multirow{2}{0.3cm}{\rotatebox{90}{1/3}}
& \texttt{HI-VAE\_piecewise} & \contour{black}{0.949±0.007} & 0.936±0.008 & \contour{black}{0.502±0.059} & 0.559±0.052 & \contour{black}{0.181±0.053} & 0.158±0.063 \\
& \texttt{HI-VAE\_weibull} & 0.937±0.008 & \contour{black}{0.941±0.008} & 0.561±0.051 & \contour{black}{0.542±0.063} & \contour{black}{0.207±0.066} & 0.152±0.062 \\
\hline
\multirow{2}{0.3cm}{\rotatebox{90}{2/3}}
& \texttt{HI-VAE\_piecewise} & \contour{black}{0.965±0.005} & 0.961±0.006 & \contour{black}{0.589±0.038} & 0.612±0.038 & \contour{black}{0.220±0.057} & 0.218±0.056 \\
& \texttt{HI-VAE\_weibull} & \contour{black}{0.953±0.006} & 0.953±0.006 & \contour{black}{0.633±0.031} & 0.637±0.035 & 0.213±0.055 & \contour{black}{0.232±0.048} \\
\hline
\multirow{2}{0.3cm}{\rotatebox{90}{3/3}}
& \texttt{HI-VAE\_piecewise} & \contour{black}{0.970±0.004} & 0.966±0.005 & \contour{black}{0.689±0.026} & 0.691±0.027 & \contour{black}{0.176±0.045} & 0.175±0.044 \\
& \texttt{HI-VAE\_weibull} & 0.953±0.005 & \contour{black}{0.958±0.004} & \contour{black}{0.710±0.026} & 0.714±0.025 & 0.213±0.047 & \contour{black}{0.214±0.043} \\
\hline
\end{tabular}
\end{center}
\end{table}

\begin{table}[!h]
\scriptsize
\begin{center}
\caption{Comparison of generative performance metrics (J-S distance, survival distance, $K$-map) across sampling strategies (posterior vs. prior) for the NCT00113763 dataset (trained on controls only).}
\begin{tabular}{cc|cc|cc|cc}
\hline
$\mathbf{\upsilon}$& \textbf{Algorithm}&\multicolumn{2}{c|}{J-S dist $\downarrow$}& \multicolumn{2}{c|}{Surv dist $\downarrow$ }& \multicolumn{2}{c}{$K$-map score $\uparrow$}  \\ [0.7ex]
& & Posterior & Prior & Posterior & Prior & Posterior & Prior  \\
\hline
\multirow{2}{0.3cm}{\rotatebox{90}{1/3}}
& \texttt{HI-VAE\_piecewise} & \contour{black}{0.007±0.001} & 0.008±0.001 & 0.039±0.017 & \contour{black}{0.027±0.013} & \contour{black}{3.385±1.279} & 2.910±1.157 \\
& \texttt{HI-VAE\_weibull} & 0.009±0.001 & \contour{black}{0.008±0.001} & \contour{black}{0.025±0.008} & 0.026±0.009 & 2.835±1.251 & \contour{black}{3.125±1.125} \\
\hline
\multirow{2}{0.3cm}{\rotatebox{90}{2/3}}
& \texttt{HI-VAE\_piecewise} & \contour{black}{0.005±0.001} & 0.007±0.001 & \contour{black}{0.025±0.01} & 0.048±0.014 & \contour{black}{10.245±1.973} & 9.455±1.93 \\
& \texttt{HI-VAE\_weibull} & \contour{black}{0.006±0.001} & 0.006±0.001 & 0.029±0.01 & \contour{black}{0.023±0.008} & \contour{black}{9.275±2.045} & 8.770±1.959 \\
\hline
\multirow{2}{0.3cm}{\rotatebox{90}{3/3}}
& \texttt{HI-VAE\_piecewise} & \contour{black}{0.004±0.001} & 0.005±0.001 & 0.026±0.009 & \contour{black}{0.017±0.007} & \contour{black}{14.425±2.847} & 13.44±2.465 \\
& \texttt{HI-VAE\_weibull} & 0.005±0.001 & \contour{black}{0.005±0.001} & 0.025±0.009 & \contour{black}{0.020±0.007} & \contour{black}{13.935±2.55} & 13.46±2.536 \\
\hline
\end{tabular}
\end{center}
\end{table}

\begin{table}[!h]
\scriptsize
\begin{center}
\caption{Comparison of generative performance metrics (KS test, Detect XGB, NNDR) across sampling strategies (posterior vs. prior) for the NCT00113763 dataset (trained on controls only).}
\begin{tabular}{cc|cc|cc|cc}
\hline
$\mathbf{\upsilon}$& \textbf{Algorithm}&\multicolumn{2}{c|}{KS test $\uparrow$}& \multicolumn{2}{c|}{Detection XGB $\downarrow$ }& \multicolumn{2}{c}{NNDR $\uparrow$}  \\ [0.7ex]
& & Posterior & Prior & Posterior & Prior & Posterior & Prior  \\
\hline
\multirow{2}{0.3cm}{\rotatebox{90}{1/3}}
& \texttt{HI-VAE\_piecewise} & \contour{black}{0.956±0.006} & 0.953±0.006 & 0.604±0.035 & \contour{black}{0.597±0.038} & \contour{black}{0.213±0.057} & 0.210±0.063 \\
& \texttt{HI-VAE\_weibull} & 0.949±0.006 & \contour{black}{0.953±0.005} & \contour{black}{0.586±0.036} & 0.599±0.036 & 0.208±0.046 & \contour{black}{0.211±0.047} \\
\hline
\multirow{2}{0.3cm}{\rotatebox{90}{2/3}}
& \texttt{HI-VAE\_piecewise} & \contour{black}{0.967±0.004} & 0.958±0.004 & \contour{black}{0.667±0.026} & 0.677±0.023 & \contour{black}{0.208±0.047} & 0.194±0.048 \\
& \texttt{HI-VAE\_weibull} & \contour{black}{0.966±0.004} & 0.964±0.004 & \contour{black}{0.664±0.024} & 0.673±0.025 & \contour{black}{0.210±0.043} & 0.204±0.037 \\
\hline
\multirow{2}{0.3cm}{\rotatebox{90}{3/3}}
& \texttt{HI-VAE\_piecewise} & \contour{black}{0.973±0.003} & 0.972±0.003 & \contour{black}{0.695±0.025} & 0.702±0.022 & \contour{black}{0.232±0.044} & 0.185±0.046 \\
& \texttt{HI-VAE\_weibull} & \contour{black}{0.971±0.003} & 0.971±0.003 & 0.719±0.022 & \contour{black}{0.713±0.021} & \contour{black}{0.216±0.035} & 0.207±0.034 \\
\hline
\end{tabular}
\end{center}
\end{table}

\begin{table}[!h]
\scriptsize
\begin{center}
\caption{Comparison of generative performance metrics (J-S distance, survival distance, $K$-map) across sampling strategies (posterior vs. prior) for the NCT00339183 dataset (trained on controls only).}
\begin{tabular}{cc|cc|cc|cc}
\hline
$\mathbf{\upsilon}$& \textbf{Algorithm}&\multicolumn{2}{c|}{J-S dist $\downarrow$}& \multicolumn{2}{c|}{Surv dist $\downarrow$ }& \multicolumn{2}{c}{$K$-map score $\uparrow$}  \\ [0.7ex]
& & Posterior & Prior & Posterior & Prior & Posterior & Prior  \\
\hline
\multirow{2}{0.3cm}{\rotatebox{90}{1/3}}
& \texttt{HI-VAE\_piecewise} & 0.007±0.001 & \contour{black}{0.007±0.001} & 0.049±0.023 & \contour{black}{0.039±0.016} & \contour{black}{7.475±2.062} & 6.595±1.926 \\
& \texttt{HI-VAE\_weibull} & 0.007±0.001 & \contour{black}{0.007±0.001} & 0.056±0.023 & \contour{black}{0.042±0.014} & \contour{black}{7.155±2.122} & 5.750±1.875 \\
\hline
\multirow{2}{0.3cm}{\rotatebox{90}{2/3}}
& \texttt{HI-VAE\_piecewise} & 0.004±0.001 & \contour{black}{0.004±0.001} & 0.025±0.01 & \contour{black}{0.023±0.009} & 2.045±0.846 & \contour{black}{2.135±0.97} \\
& \texttt{HI-VAE\_weibull} & \contour{black}{0.005±0.001} & 0.005±0.001 & \contour{black}{0.038±0.013} & 0.044±0.013 & 2.165±1.016 & \contour{black}{2.220±0.993} \\
\hline
\multirow{2}{0.3cm}{\rotatebox{90}{3/3}}
& \texttt{HI-VAE\_piecewise} & 0.004±0.001 & \contour{black}{0.004±0.001} & 0.022±0.009 & \contour{black}{0.022±0.009} & 2.720±1.13 & \contour{black}{3.000±1.211} \\
& \texttt{HI-VAE\_weibull} & 0.005±0.001 & \contour{black}{0.004±0.001} & \contour{black}{0.034±0.01} & 0.038±0.011 & 2.490±1.173 & \contour{black}{2.600±1.199} \\
\hline
\end{tabular}
\end{center}
\end{table}

\begin{table}[!h]
\scriptsize
\begin{center}
\caption{Comparison of generative performance metrics (KS test, Detect XGB, NNDR) across sampling strategies (posterior vs. prior) for the NCT00339183 dataset (trained on controls only).}
\begin{tabular}{cc|cc|cc|cc}
\hline
$\mathbf{\upsilon}$& \textbf{Algorithm}&\multicolumn{2}{c|}{KS test $\uparrow$}& \multicolumn{2}{c|}{Detection XGB $\downarrow$ }& \multicolumn{2}{c}{NNDR $\uparrow$}  \\ [0.7ex]
& & Posterior & Prior & Posterior & Prior & Posterior & Prior  \\
\hline
\multirow{2}{0.3cm}{\rotatebox{90}{1/3}}
& \texttt{HI-VAE\_piecewise} & 0.955±0.007 & \contour{black}{0.958±0.007} & \contour{black}{0.519±0.05} & 0.522±0.05 & \contour{black}{0.189±0.055} & 0.173±0.053 \\
& \texttt{HI-VAE\_weibull} & 0.957±0.007 & \contour{black}{0.958±0.007} & 0.526±0.046 & \contour{black}{0.517±0.05} & \contour{black}{0.171±0.057} & 0.161±0.052 \\
\hline
\multirow{2}{0.3cm}{\rotatebox{90}{2/3}}
& \texttt{HI-VAE\_piecewise} & 0.969±0.005 & \contour{black}{0.971±0.004} & \contour{black}{0.658±0.034} & 0.667±0.034 & \contour{black}{0.127±0.037} & 0.122±0.035 \\
& \texttt{HI-VAE\_weibull} & \contour{black}{0.967±0.005} & 0.966±0.005 & \contour{black}{0.676±0.037} & 0.677±0.037 & 0.124±0.047 & \contour{black}{0.131±0.039} \\
\hline
\multirow{2}{0.3cm}{\rotatebox{90}{3/3}}
& \texttt{HI-VAE\_piecewise} & \contour{black}{0.973±0.004} & 0.973±0.004 & 0.805±0.022 & \contour{black}{0.803±0.023} & 0.135±0.039 & \contour{black}{0.147±0.037} \\
& \texttt{HI-VAE\_weibull} & 0.969±0.004 & \contour{black}{0.973±0.004} & \contour{black}{0.803±0.026} & 0.809±0.023 & 0.125±0.037 & \contour{black}{0.130±0.04} \\
\hline
\end{tabular}
\end{center}
\end{table}

\clearpage

\subsection{Preliminary exploration of differential privacy} \label{subsec:differential_privacy}

We present here our preliminary results exploring differential privacy in our models, using the \texttt{Opacus} framework~\citep{opacus} (Section~\ref{sec:extended_XP} of the main paper). Specifically, we applied the privacy engine to our HI-VAE model with the following parameters:
\begin{verbatim}
hivae_model, optimizer, data_loader = privacy_engine.make_private(
        module=hivae_model,
        optimizer=optimizer,
        data_loader=data_loader,
        noise_multiplier=2.0,
        max_grad_norm=1.0)
\end{verbatim}

\begin{figure}[h!]
    \centering
    \includegraphics[width=0.6\columnwidth]{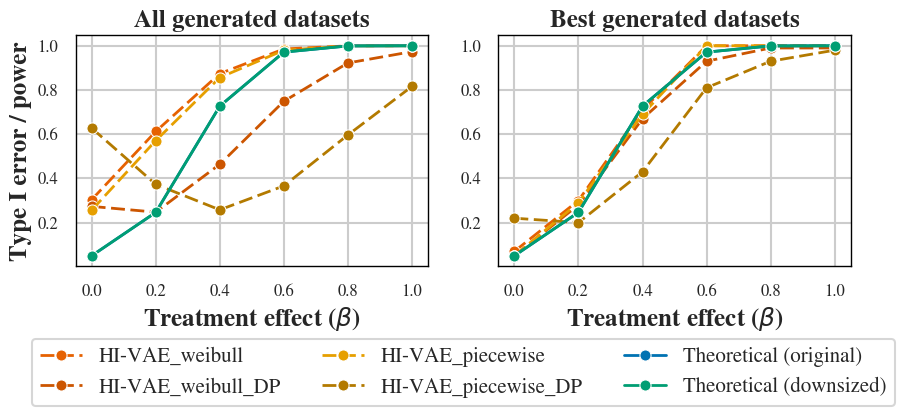}
    \caption{\footnotesize Type I error and power estimation in the independent simulation setting, (trained on controls only, replacement case), with or without \textbf{the differential privacy method}. Dashed lines: empirical power. Green: theoretical power with reduced control size. Blue: theoretical power with generated control size.}
    \label{fig:simu_pvalue_comp_DP}
\end{figure}

\begin{figure}[h!]
    \centering
    \includegraphics[width=0.88\columnwidth]{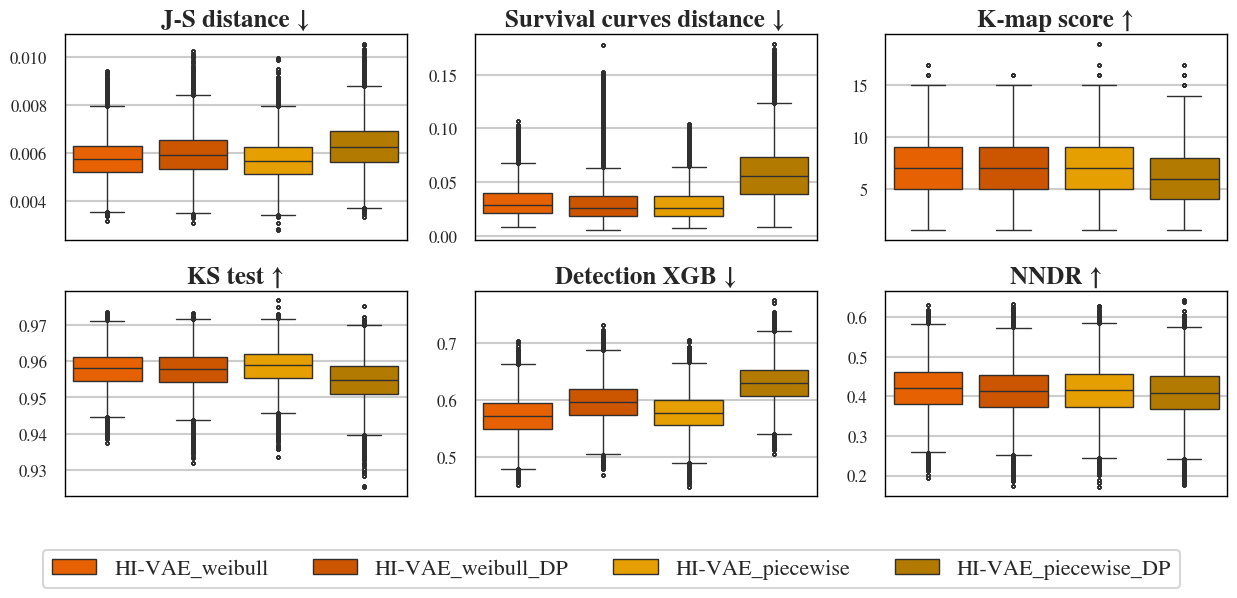}
    \caption{\footnotesize Comparison of generative performance metrics (J-S distance, survival distance, $K$-map)  in the independent simulation setting, (trained on controls only, replacement case), with or without \textbf{the differential privacy method}. Dashed lines: empirical power. Green: theoretical power with reduced control size. Blue: theoretical power with generated control size.}
    \label{fig:simu_perf_comp_DP}
\end{figure}

\end{document}